%% file: main-article.tex
\documentclass[10pt, a4paper, twocolumn]{article}
\usepackage{times}
\usepackage{authblk}
\usepackage{graphicx}%
\usepackage{multirow}%
\usepackage{amsmath,amssymb,amsfonts}%
\usepackage{amsthm}%
\usepackage{mathrsfs}%
\usepackage[title]{appendix}%
\usepackage{xcolor}%
\usepackage{algpseudocode}
\usepackage{textcomp}%
\usepackage{manyfoot}%
\usepackage{booktabs}%
\usepackage{listings}%
\usepackage{microtype}
\usepackage{graphicx}
\usepackage{wrapfig}
\usepackage{float}
\usepackage{comment}
\usepackage{xcolor}
\usepackage{caption, subcaption}
\usepackage{textcase}
\usepackage{multirow}
\usepackage{booktabs}
\usepackage{colortbl}
\usepackage{lipsum}
\usepackage{lineno}

\usepackage[ruled, noline, commentsnumbered]{algorithm2e}
\SetKwFor{For}{for (}{) $\lbrace$}{$\rbrace$}
\SetNlSty{textbf}{}{\hspace{-4em}\hspace{3em}}
\SetAlgoNlRelativeSize{-1} 

\usepackage[maxbibnames=99, backend=biber,style=nature]{biblatex}
\definecolor{customgreen}{RGB}{ 5, 102, 8}
\definecolor{custompink}{RGB}{ 255, 0, 127}
\definecolor{customviolet}{RGB}{72, 99, 160}
\definecolor{table_color}{RGB}{255, 203, 164} 
\definecolor{table_color_1}{RGB}{224, 255, 255}

\usepackage{tikz}
\newcommand*\circled[1]{\tikz[baseline=(char.base)]{
            \node[shape=circle,draw,inner sep=1pt] (char) {#1};}}

\newcommand{\ours}{MetaWearS}
\usepackage{xcolor,colortbl}

\definecolor{Gray}{gray}{0.85}
\definecolor{LightCyan}{rgb}{0.88,1,1}

\usepackage{tcolorbox}

\usepackage{geometry}

\newcommand{\abstractText}{\noindent

Wearable systems provide continuous health monitoring and can lead to early detection of potential health issues. However, the lifecycle of wearable systems faces several challenges. First, effective model training for new wearable devices requires substantial labeled data from various subjects collected directly by the wearable. Second, subsequent model updates require further extensive labeled data for retraining. Finally, frequent model updating on the wearable device can decrease the battery life in long-term data monitoring. Addressing these challenges, in this paper, we propose MetaWearS, a meta-learning method to reduce the amount of initial data collection required. Moreover, our approach incorporates a prototypical updating mechanism, simplifying the update process by modifying the class prototype rather than retraining the entire model. We explore the performance of MetaWearS in two case studies, namely, the detection of epileptic seizures and the detection of atrial fibrillation. We show that by fine-tuning with just a few samples, we achieve 70\% and 82\% AUC for the detection of epileptic seizures and the detection of atrial fibrillation, respectively. Compared to a conventional approach, our proposed method performs better with up to 45\% AUC. Furthermore, updating the model with only 16 minutes of additional labeled data increases the AUC by up to 5.3\%. Finally, MetaWearS reduces the energy consumption for model updates by 456x and 418x for epileptic seizure and AF detection, respectively.
}

\begin{document}
\title{MetaWearS: A Shortcut in Wearable Systems Lifecycle with Only a Few Shots}

\author[1]{Alireza Amirshahi$^{\dagger,}$}
\author[2]{Maedeh H. Toosi$^{\dagger,}$}
\author[2, 3]{Siamak Mohammadi}
\author[1]{Stefano Albini}
\author[1]{Pasquale Davide Schiavone}
\author[1]{Giovanni Ansaloni}
\author[4]{Amir Aminifar}
\author[1]{David Atienza}
\affil[1]{Embedded Systems Laboratory (ESL), École Polytechnique Fédérale de Lausanne (EPFL), Switzerland}
\affil[2]{School of Electrical and Computer Engineering, University of Tehran, Iran}
\affil[3]{School of Computing Science, Institute for Research in Fundamental Sciences (IPM), Tehran, Iran}
\affil[4]{Department of Electrical and Information Technology, Lund University, Sweden}
\affil[1]{\textit {Email: \{alireza.amirshahi, stefano.albini, davide.schiavone, giovanni.ansaloni, david.atienza\}@epfl.ch}, }
\affil[2]{\textit {\{maedeh.toosi,smohamadi\}@ut.ac.ir}, }
\affil[3]{\textit {amir.aminifar@eit.lth.se}, }
\affil[ ]{\textit{\small $^\dagger$ These authors contributed equally to this work.}}
\renewcommand\Affilfont{\itshape\small}
\date{}                     %% if you don't need date to appear

\twocolumn[
  \begin{@twocolumnfalse}
    \maketitle
    \begin{abstract}
      \abstractText
      \newline
      \newline
    \end{abstract}
  \end{@twocolumnfalse}
]

\input{sections/introduction}
\input{sections/Background}
\input{sections/methods}

\input{sections/experimental_setup}
\input{sections/result}

\input{sections/discussion}

\section{CONCLUSION}\label{conclusion}

In this paper, we have proposed a novel few-shot learning method to detect abnormalities in scarce signals acquired by wearable systems. In the first phase of the wearable device's lifecycle, we have significantly reduced the amount of data collection for the initial training of the network. We have shown that by using only three subjects in the epileptic seizure detection use case, we can fine-tune the model and obtain 86.8\% of AUC. This performance is 82\% for AF detection with only 5 minutes of data. In the second phase of the wearable's lifecycle, where the wearable systems are being used by the end-users, we have also reduced the required labeled data for updating the framework using \ours. We have shown that with only 16 minutes of additional labeled signals, we can improve AUC by 75.6\% in the epilepsy task. Furthermore, we have measured the time and power consumption for the transformer model for a 12-second signal window as 1.9 seconds and 50~mW, respectively. The update process, due to \ours, can consume up to 456 times less than updating the entire model.

% \section{DATA AVAILABILITY}
% The datasets utilized for seizure detection, including both training and testing data, are publicly available. For the Atrial Fibrillation case, the training data is publicly available. However, the test dataset is part of a proprietary dataset provided by a private collaborator. According to the signed data usage agreement with the data owners, the data cannot be published. Access to this dataset is restricted to the terms and limitations established by the data owner. More information about this dataset and access requests can be directed to the corresponding author.%them through ZivaTec's \href{https://zivatec.com/}{website}.

% \section{CODE AVAILABILITY}
% The code for the experiments is available in the following repository: \\
% \href{https://github.com/alirezaamir/MetaWearS}{https://github.com/alirezaamir/MetaWearS}
%%===========================================================================================%%
%% If you are submitting to one of the Nature Portfolio journals, using the eJP submission   %%
%% system, please include the references within the manuscript file itself. You may do this  %%
%% by copying the reference list from your .bbl file, paste it into the main manuscript .tex %%
%% file, and delete the associated \verb+\bibliography+ commands.                            %%
%%===========================================================================================%%

%\bibliography{sn-bibliography}% common bib file
%% if required, the content of .bbl file can be included here once bbl is generated
%%\input sn-article.bbl
\printbibliography

\end{document}

%% file: sections/introduction.tex
\section{Introduction}
\label{sec:intro}
Advancements in biomedical signal monitoring techniques have contributed significantly to the early diagnosis and treatment of various medical conditions. 
Immediate and accurate detection of these conditions can guide appropriate medical interventions, thereby improving patient care{\cite{rangayyan2024biomedical,kwong2011energy}}. 

Wearable systems are increasingly being used to monitor health conditions and provide real-time data on various physiological parameters. Wearable devices, such as smartwatches, fitness trackers, and specialized medical wearables, can track and record data related to heart rate, blood pressure, sleep patterns, physical activity, etc.{\cite{chiauzzi2015patient,patel2012review,chan2012smart}}.

\begin{figure}[t]
    \centering
    \includegraphics[width=1\linewidth, height=5cm]{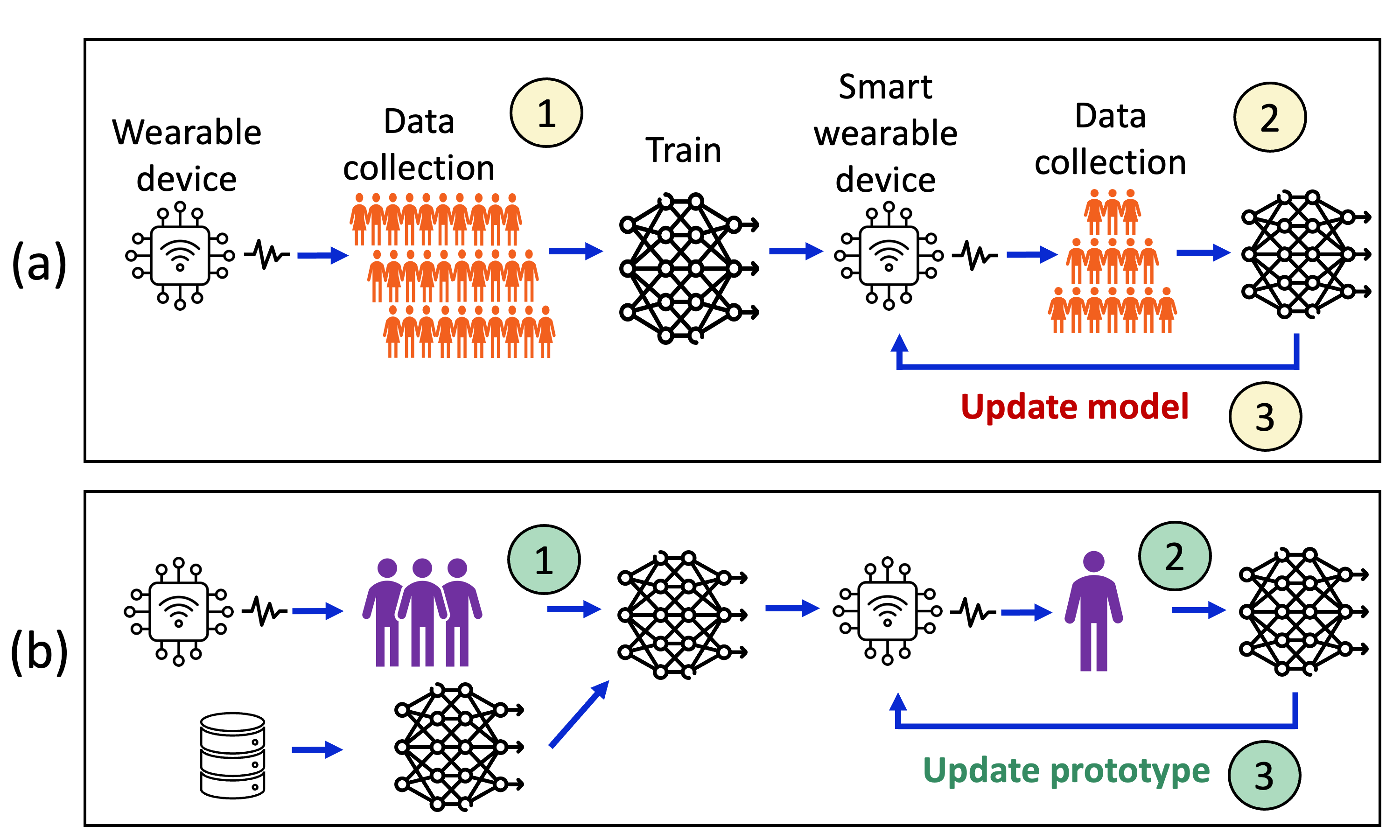}
    \caption{(a) Challenges in a wearable system lifecycle: \protect\circled{1} Initial data collection, \protect\circled{2} further data collection for updates, and \protect\circled{3} inefficient update protocol. (b) The proposed method addresses all the mentioned challenges.}
    \label{fig:overview}
\end{figure}

Deep learning has demonstrated significant advances in a variety of biomedical applications. However, implementing these advances into practical wearable systems often faces numerous challenges. These challenges span the entire lifecycle of a wearable device, from its data collection phase to its utilization by end-users. Fig.~\ref{fig:overview}a shows the definition of a wearable system lifecycle and its challenges. The first challenge, shown in Fig.~\ref{fig:overview}a with \circled{1}, is the requirement for substantial biomedical labeled data to train deep learning models.  Data collection is a costly and time-consuming process{\cite{zakim2015data}}. This issue is further exacerbated in wearable systems by the fact that the data beneficial to the model training should ideally be collected from identical devices{\cite{lasko2024probabilistic, bachtiger2022point, vijayan2021review, dexter2020generalization}}. 

%Furthermore, managing model updates in wearables necessitates efficient processing due to limited resources, careful storage, and memory management during updates.

% Once data are collected from numerous subjects, the deep learning model is trained, enabling wearable devices to become a smart system capable of simultaneous data collection and abnormality detection. This marks the phase where the wearable device is utilized by the users. 
The next challenge for deep learning deployment in wearable systems is the data requirement for the updating process, shown by \circled{2} in Fig.~\ref{fig:overview}a. In this phase, data should ideally be collected from a large number of new participants to have enough data to retrain the model. 
Otherwise, with a limited number of data, the model can suffer from an overfitting problem, where it becomes too fit to the new training data and performs poorly on unseen data{\cite{ying2019overview,hawkins2004problem}}.
% Note that model retraining is typically performed on a more powerful computing device, such as a server.
%

Finally, once new signals are obtained and the model is updated, the updated model should be sent back and replaced on the wearable system. However, deep learning models are typically large. When models are transmitted and read by processors from a wireless network, modeling transmission can be quite time-consuming and energy-intensive{\cite{chen2019deep}}. This significantly impacts battery life, a critical factor for wearable devices.  As depicted in Fig.~\ref{fig:overview}a by \circled{3}, the third challenge is a high energy consumption for frequent model updates.
% received from the server to ensure the performance of the wearable system.

In our paper, we introduce \ours, a novel \underline{Meta}-learning method for \underline{Wear}able \underline{S}ystems that addresses the critical challenges of data scarcity and resource efficiency. Remarkably, as shown in Fig.~\ref{fig:overview}b, our approach reduces the data required for the initial training of the model (\circled{1}) and for the update process (\circled{2}). This is achieved through a targeted modification of the few-shot learning strategy, specifically designed to address the aforementioned challenges.

Furthermore, our method incorporates an energy-efficient updating mechanism, and instead of retraining and transmitting the entire model, we update only a single vector, called \emph{prototype}. In line with the principles of prototypical networks{~\cite{snell2017prototypical}}, the prototype size is considerably smaller than the model weight. Therefore, the processor consumes less time and energy to receive the prototypes instead of receiving the whole model (\circled{3}). This can significantly increase the battery life time.

To evaluate \ours, Epilepsy and Atrial Fibrillation~(AF) arrhythmia, two distinct but critical biomedical conditions, are chosen as two case studies based on Electroencephalogram~(EEG) and Electrocardiogram~(ECG), respectively, as two different types of biomedical signals. Moreover, we use two types of deep learning models for our study: MobileNetV2{~\cite{sandler2018mobilenetv2}} and VisionTransformer{~\cite{dosovitskiy2020image}}.
% due to their prevalence and the inherent challenges they pose in detection. 
%The objective of this evaluation is to show the method's capacity across varying domains and tasks. 
This not only highlights the generality and adaptability of our approach, but also emphasizes its potential to surpass the limitations of traditional single-domain evaluations in biomedical signals.

% To assess the practicality and real-world applicability of our approach, we implement the proposed model on X-HEEP as an energy-efficient and low-power open-source hardware platform{~\cite{schiavone2023x}}. This hardware evaluation aims to determine the model's suitability for integration into wearable devices, where efficiency and real-time performance are important.
% This paper presents three key contributions in the application-based hardware design for AF arrhythmia detection as follows:

% Epilepsy, a neurological brain disorder characterized by recurrent seizures, demands precise and rapid identification to aid in tailored treatment strategies~\cite{alarcon2012introduction}. Similarly, AF, the most common type of irregular heartbeat, is a serious cardiac condition linked to complications such as stroke. Heart disease remains the leading global cause of death, with deaths due to heart disease and stroke on the rise since 2000. Prompt AF diagnosis is crucial to prevent complications~\cite{chugh2014worldwide}. 

This paper presents four key contributions in the application-based wearable system design for biomedical abnormalities monitoring as follows:

\begin{itemize}

\item Our work introduces \ours{} as a few-shot learning strategy specifically designed for the effective detection of abnormalities in biomedical signals, particularly in scenarios where initial labeled data is scarce. 

\item The proposed \ours{} can update the models with minimal additional labeled data when end users are actively using the wearable system. This demonstrates the practical advantages of our methodology in reducing the data requirement.

% This update process involves modifying only the prototypes without necessitating complete retraining and requires a minimal number of newly labeled signals. By utilizing our meta-transfer learning approach, we effectively reduce the volume of newly annotated samples needed to update the model, demonstrating the practical advantages of our methodology in reducing the data.

\item We propose an update process that involves modifying only the prototypes without necessitating retraining and replacing the entire model. This update protocol significantly improves battery life.

\item To assess the generalization capabilities of \ours{} in the context of few-shot learning, we carried out comprehensive experiments on two separate case studies with two types of biomedical signals: AF arrhythmia detection using ECG and epileptic seizure detection using EEG signals.

\end{itemize}

%% file: sections/Background.tex
\section{Background}
\label{sec:Background}
\begin{figure}[t]
\centering
\begin{subfigure}{\linewidth}
    \centering
    \includegraphics[width=0.9\linewidth]{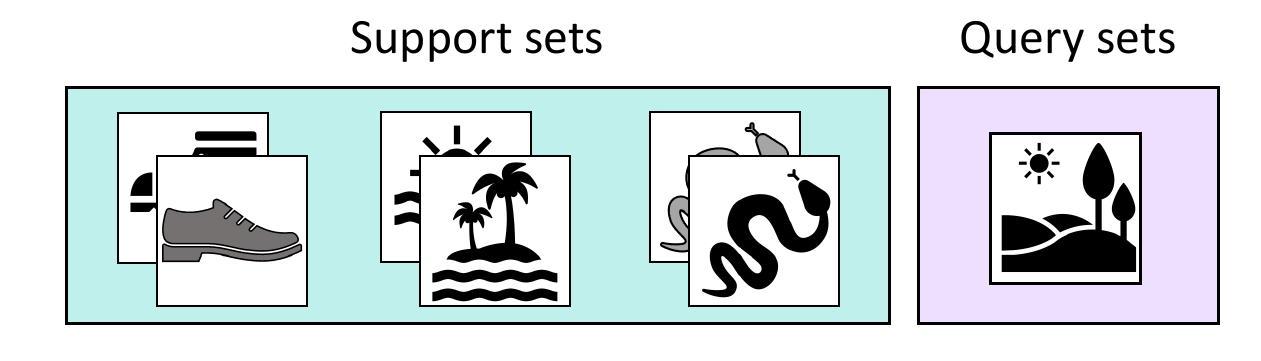}
    \subcaption{Meta-testing}
    \label{fig:few_shot_meta_test}
\end{subfigure}

\begin{subfigure}{\linewidth}
    \centering
    \includegraphics[width=0.9\linewidth]{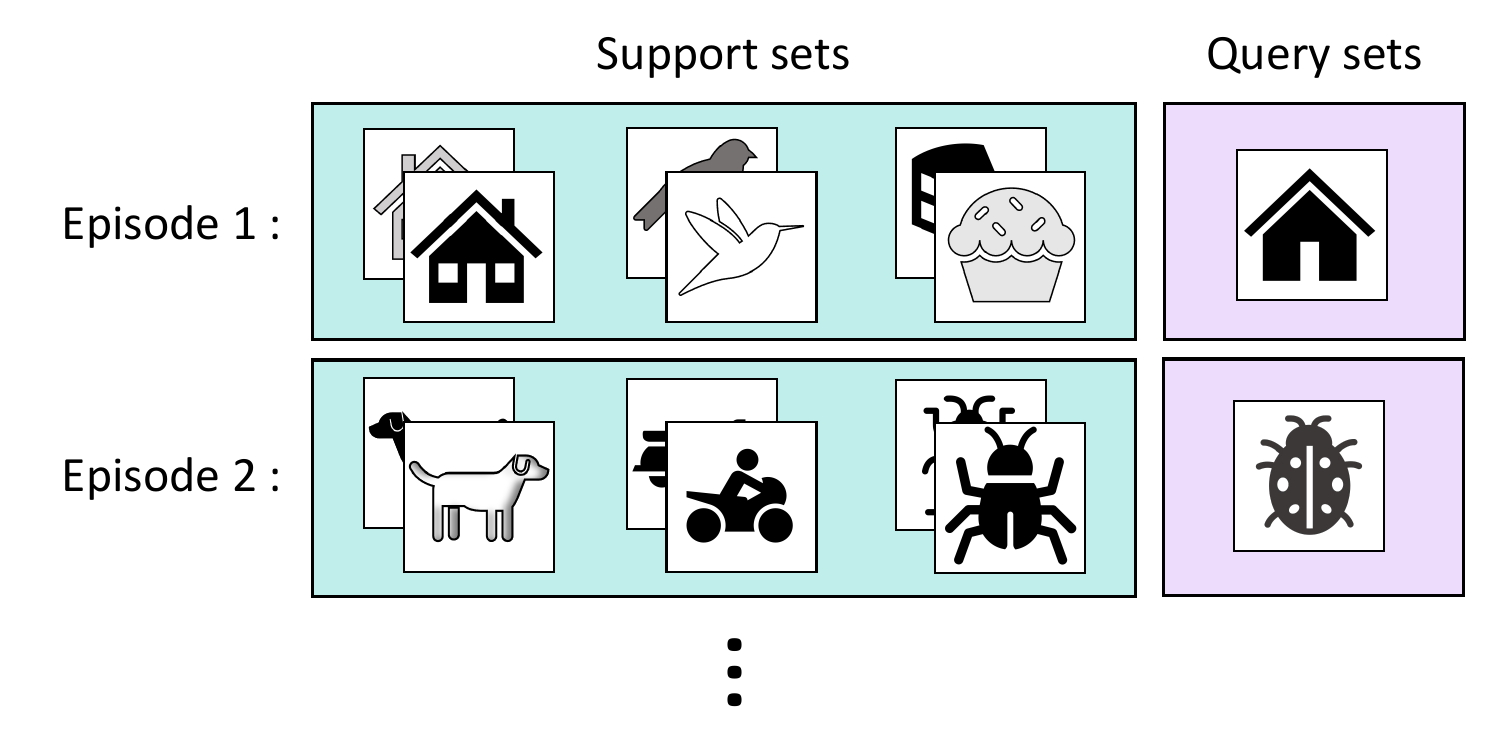}
    \subcaption{Meta-training}
    \label{fig:few_shot_meta_train}
\end{subfigure}
\caption{Meta-testing and Meta-training in few-shot learning, highlighting the creation of episodes with three distinct classes and two samples randomly selected from a base dataset. The meta-testing process involves entirely new classes to form a support set and query set, mirroring the tasks in the meta-training phase. }
\end{figure}

\textbf{Few-shot learning} methods aim at designing a model that classifies a test sample based on a very limited number of previously seen labeled examples. 
For instance, Fig.~\ref{fig:few_shot_meta_test} shows an example of a few-shot learning \emph{task}. In this figure, the test sample must be classified only using a small group of limited labeled samples, referred to as \emph{support set}. 

A potential solution for the few-shot learning challenge, in the context of meta-learning, is to simulate the situation of having a few labeled samples during training. Essentially, the model is trained on a large dataset, referred to as the \emph{base dataset}, to learn the similarities and differences among a few objects. This step of training the model on a base dataset is called \emph{meta-train}, shown in Fig.~\ref{fig:few_shot_meta_train}. 

The training step, as shown in Fig.~\ref{fig:few_shot_meta_train}, involves several \emph{episodes}. In each episode, $N$ distinct classes are randomly selected from the base dataset, and  $k$ random samples are chosen from each class. These samples collectively form a support set for the episode.

Let $\mathcal{D} ^{\text{train}}$ be the base dataset for meta-training. The support set for episode $e$ is randomly selected from $\mathcal{D} ^{\text{train}}$ to create $S_e = \left\{\left(x_i^s, y_i^s\right)\right\}_{i=1}^{k \cdot N}$.
The goal of this episode is to classify new samples $x_j$ that belong to these N classes. Thus, we create a query set with these new samples as $Q_e =  \left\{\left(x_j^q, y_j^q\right)\right\}_{j=1}^{m \cdot N}$, where $m$ is the number of new samples in each class.  The loss function during meta-training aims to minimize the log-likelihood of each predicted class $\hat{y}_j^q$ of the samples in $Q_e$ compared to the corresponding ground truth $y_j^q$. 

As depicted in Fig.~\ref{fig:few_shot_meta_train}, the process of creating episodes and selecting samples and classes is repeated multiple times during meta-training. The underlying idea behind meta-training is to simulate a situation where we have a few labeled examples in the support set, and the model needs to classify the class of a new test sample in the query set. 

In few-shot learning, a \emph{meta-test} is created to evaluate the model. This involves entirely new classes that were not present in $\mathcal{D} ^{\text{train}}$, forming  $\mathcal{D} ^{\text{test}}$, also referred to as the \emph{target dataset}. As shown in Fig.~\ref{fig:few_shot_meta_test}, we again have a support set with the same number of classes and samples, and the objective is to classify the samples in the query set. This task mirrors the tasks in the meta-train.

\textbf{Prototypical Networks}{~\cite{snell2017prototypical}} are a type of algorithm designed for the few-shot learning challenge. In this approach, a machine learning or deep learning model computes a feature vector, denoted as $f_\phi(\mathbf{x}_i)$, for each sample $\mathbf{x}_i$ in the support set $S_e$, where $\phi$ represents the model's parameters. This feature vector $f_\phi(\mathbf{x}_i)$, in the context of deep learning, is usually the output of the layer that precedes the final output layer. 

In each episode, prototypes $\mathbf{c}_n$ are calculated for each class $n$ within the support set $S^n$ as the average of the feature vectors of all instances belonging to their corresponding classes. The formula for calculating the prototypes is as follows: 
\begin{equation}
\mathbf{c}_n=\frac{1}{\left|S^n\right|} \sum_{\left(\mathbf{x}_i, y_i\right) \in S^n} f_\phi\left(\mathbf{x}_i\right),
\label{eq:prototypes}
\end{equation}

The query set is also fed into the model, and a feature vector $f_\phi(\mathbf{x}_j)$ is extracted for each sample $\mathbf{x}_j$. The classification of the query set is then performed by comparing the feature vectors of the query set with the class prototypes $\mathbf{c}_n$. We can classify the query set by identifying the nearest prototype in this embedding space to the query samples.

%% file: sections/methods.tex
\section{PROPOSED METHODS}
\label{sec:method}

This section is structured into the following steps. Initially, we explain how \ours{} tackles challenge \circled{1}. As mentioned in Section~\ref{sec:intro}, this challenge pertains to the initial data collection required for training the first model.
In Section \ref{sec:method:phase2}, we employ our proposed few-shot learning method to address Challenge~\circled{2}, aiming to reduce the demand for a large quantity of annotated new samples for updates during their usage by end-users.
Finally, in Section~\ref{sec:method:xheep}, we discuss the solution for Challenge~\circled{3}. In this section, we explain the impact of  \ours{} on the efficient updating of prototypes and models on the hardware to minimize energy and time consumption.

\subsection{\ours{} for reducing initial data collection}\label{sec:method:phase1}

% \begin{enumerate}    
%     \item What is the definition of a task in this scenario? What are the differences between task $\tau_i$ and $\tau_j$?
%     \item How do we use Meta-transfer learning?  Base Learner, Fine-tune, Test 
%     \item What is the loss function? How do we update the weight?
%     \item What are the support set and query set when training the base learner? 
%     \item What are the support set and query set when \textbf{fine-tuning} the new dataset? 
%     \item What is the scenario after fine-tuning a few samples we collected in the first Phase? How is the model deployed on a wearable device and used in tests?
% \end{enumerate}

Our proposed method is inspired by the meta-transfer learning method{~\cite{sun2019meta}}. Meta-transfer learning is a general approach in which pre-trained models on large datasets are used as a foundation for new tasks, often referred to as fine-tuning. The goal of transfer learning, in this method, is to use existing knowledge and reduce the need for extensive new training data. 

 Algorithm~\ref{alg:myalgorithm} formulates the explained meta-train procedure applied initially to the base dataset for pretraining and, consequently, to the target dataset for fine-tuning. During meta-train, each episode consists of a support set and a query set. The support set contains a small number $k$ of samples from each class (normal and abnormal) from $N_S$ patients. Similarly, the query set contains $k$ samples of the same classes from $N_Q$ patients. Based on prototypical networks, in our proposed method, both support and query set samples are processed by the model to extract features $f_\phi(\mathbf{x}_i) \in \mathbb{R}^D$ within each episode. $f_\phi(\mathbf{x}_i)$ is considered as a $D$-dimensional vector extracted from the output of the layer that precedes the final output layer in the models. 
 % The model is pretrained to extract discriminative features and generalize them to the base dataset classes. Then, the model is fine-tuned on the target dataset using the knowledge and features learned from the base dataset. 
 %
 %The support set $S=$ $\left\{\left(\mathbf{x}_1, y_1\right), \ldots,\left(\mathbf{x}_N, y_N\right)\right\}$ has $N$ labeled samples and $f_\phi(\mathbf{x}_i) \in \mathbb{R}^D$ is the $D$-dimensional feature vector and $y_i \in\{1, \ldots, K\}$ is the corresponding label.
\begin{algorithm}[t]
\Indp
    \SetKwInOut{header}{\textcolor{customviolet}{\textit{Phase}}}
    \SetKwInOut{KwIn}{Input}
    \SetKwInOut{KwOut}{Output}
    \LinesNumbered
    % \header{\textcolor{customviolet}{\bfseries \textit{ Meta Training}}}
    \KwIn{ $f_\phi$, $\mathcal{D}^{\text{train}}$}
    \KwOut{Updated loss $\mathcal{L}_{\mathcal{D}}$}
   $ Pat_S \leftarrow \text { Random Sample }\left(\text{Patients}, N_S\right)$ 
   
    $ Pat_Q \leftarrow \text { Random Sample }\left(\text{Patients} \backslash Pat_S, N_Q\right)$

   \For{$n$ in $\{\text{normal},\, \text{abnormal}\} $}{
        $ S^n \leftarrow \text { Random Sample}\left(\mathcal{D}^{\text{train}, n}_{Pat_S}, k\right)$;

        $ Q^n \leftarrow \text{Random Sample}\left(\mathcal{D}^{\text{train}, n}_{Pat_Q}, k\right)$ ;
        
        $\text{Compute prototypes $c_n$ from $S^n$ in \eqref{eq:prototypes}}$; 

    }
    
    \For{$(x,y)$ in  \(Q\)}{
     $\text{Update classification loss $\mathcal{L}_{\mathcal{D}}$ by \eqref{eq:3}}$ ;
     
     $\text{Optimize $\phi$}$;
   }
    \caption{A meta-train episode in \ours{} for base and target dataset}
    \label{alg:myalgorithm}
\end{algorithm}

As discussed in Section~\ref{sec:Background}, a class prototype $c_n$ for class $n$ is defined as the average of all vectors $f_\phi(\mathbf{x}_i)$ in the support set such that ${y}_i = n$. After computing the prototypes in the query dataset, classes with prototypes that are closer to the query feature $f_\phi(\mathbf{x}_i)$ obtain higher probability scores.
The similarity between the query sample and each class prototype is computed using squared Euclidean distance. Learning proceeds by applying softmax over the negative of these distances and the probability for each class is obtained. The loss function $\mathcal{L}$  is computed as Equation \eqref{eq:3}.

% \begin{equation}
% \phi =:\phi -\alpha \nabla \mathcal{L}_{\mathcal{D}}(\phi),
% \label{eq:2}
% \end{equation}
% where $\alpha$ denotes the learning rate and $\mathcal{L}$ denotes the following loss:
\begin{equation}
\mathcal{L}_{\mathcal{D}}(\phi)=\frac{1}{|\mathcal{D}|} \sum_{(\mathbf{x}, y) \in \mathcal{D}} l\left( \sigma(-\left\|f_\phi(\mathbf{x})-c\right\|_2^2), y\right),
\label{eq:3}
\end{equation}
where $l$ is the log-likelihood function and $\sigma$ is softmax.

In the inference meta-test step, the prototypes are computed using Equation \eqref{eq:prototypes}. Subsequently, the fine-tuned model, along with the prototypes, is deployed to the wearable system for patient use. 

Algorithm~\ref{alg:phase1} provides a detailed explanation of inference in wearable systems. 
When a patient uses the wearable, the acquired signal $\mathbf{x}_j$ is processed through the network, resulting in $f_\phi({\mathbf{x}_j})$.  The next step involves finding the Euclidean distance between this feature vector and the prototypes. Consequently, we can obtain the probability score for each class and classify $\mathbf{x}_j$.

\begin{algorithm}[t]
\Indp
    \setcounter{AlgoLine}{0}
    \SetKwInOut{header}{\textcolor{customviolet}{\textit{Phase}}}
    \SetKwInOut{KwIn}{Input}
    \SetKwInOut{KwOut}{Output}
    \LinesNumbered
    % \header{\textcolor{customviolet}{\bfseries \textit{ Meta Test}}}
    \KwIn{ $f_\phi$ ,
       $\mathcal{D}^{\text{train}}$, $\mathbf{x}_{j}$ in $\mathcal{D}^{\text{test}}$}
    \KwOut{Sample's probability scores}
    $ Pat_S \leftarrow \text { Random Sample }\left(\text{Patients}, N_S\right)$ 
    
    \For{$n$ in $\{\text{normal},\, \text{abnormal}\} $}{
        $ S^n \leftarrow$\;Random Sample ($\mathcal{D}^{\text{train}, n}_{Pat_S}$, $k$); \label{line:s_n}

        $\text{Compute class prototypes $c_n$ from $S^n$ in \eqref{eq:prototypes}}$;
    }

    Compute Euclidean distances: $\left\|f_\phi(\mathbf{x}_j)-\mathbf{c}\right\|_2^2 $;

    Compute the probability scores $p(\mathbf{x}_j | y)$
    
    \caption{Inference of $\mathbf{x}_j$ in the test set to tackle Challenge \protect\circled{1}}
    \label{alg:phase1}
\end{algorithm}

\subsection{\ours{} for updating the model using few new shots}\label{sec:method:phase2}
In this section, we discuss the methods corresponding to Challenge~\circled{2}.
Assume that some new annotated samples are collected from new subjects. We denote the new dataset from these subjects as $\mathcal{D}^{\text{new}}$.
%, with each sample represented as  ($\mathbf{x}^{\text{new}}_i$, $y^{\text{new}}_i$)  $\in$ $\mathcal{D}^{\text{new}}$. 
By feeding the samples of $\mathcal{D}^{\text{new}}$ into the model, we are able to compute $f_\phi (\mathbf{x})$ for all the samples. As discussed in Section~\ref{sec:Background}, we use the samples in the target set ($\mathcal{D}^{\text{train}}$) and the small sample set in $\mathcal{D}^{\text{new}}$ to reconstruct the support set for each class $n$ as follows:

\begin{align}
   \quad S^n \leftarrow & \text {Random Sample}\left(\mathcal{D}^{\text{train}, n}_{Pat_S}, k\right) \cup \nonumber \\
                        & \text {Random Sample}\left(\mathcal{D}^{\text{new}, n}, k\right) \label{eq:updated_sk} ,           
\end{align}

% \begin{algorithm}[t]
% %\SetNlSty{textbf}{(}{)}
% \Indp
%     \setcounter{AlgoLine}{0}
%     \SetKwInOut{header}{\textcolor{customviolet}{\textit{Phase}}}
%     \SetKwInOut{KwIn}{Input}
%     \SetKwInOut{KwOut}{Output}
%     \LinesNumbered
%     \KwIn{ $f_\theta$ ,
%        $\mathcal{D}^{\text{train}}$, $\mathcal{D}^{\text{new}}$, $\mathbf{x}_{i}$ in $\mathcal{D}^{\text{test}}$}
%     \KwOut{Sample's probability scores}
%     \For{$k$ in $\left\{1, \ldots, K\right\} $}{
%         $ S_k \leftarrow$\; Reconstruct the support set with \eqref{eq:updated_sk};
        
%         $\text{Compute class prototypes $c_k$ from $S_k$ in \eqref{eq:prototypes}}$;
        
%         $\text{Compute Euclidean distances: } \left\|f(\mathbf{x}_j)-c_k\right\|_2^2 $;
%     }

%     Compute the probability scores $p(\mathbf{x}_i | y)$
    
%     \caption{Inference for phase 2 in the target dataset}
%     \label{alg:phase2}
% \end{algorithm}

where $k \leq |\mathcal{D}^{\text{new}}| \ll  |\mathcal{D}^{\text{train}}|$. 
% Note that since we have already fine-tuned the model on $\mathcal{D}^{\text{train}}$, using the samples in this set has no cost. Therefore, we add these samples to the new support set. 
The inference algorithm in this part is identical to Algorithm~\ref{alg:phase1} except Line~\ref{line:s_n}, where $S^n$ should be reconstructed by the support sets based on the assignment \eqref{eq:updated_sk}. For each $(\mathbf{x}_j, y_j) \in \mathcal{D}^{\text{test}}$, $f_\phi(\mathbf{x}_j)$ is compared to the updated prototypes $c_n$ to compute the Euclidean distance and classify the sample.

In this phase, we calculate the prototypes based on the few new shots that have been received and annotated, typically on the server side. Once these new prototypes are updated, they can be transmitted back to the wearable systems. By replacing and updating these prototypes in wearable systems, subjects can benefit from a more generalized prototype, leading to improved performance. 

These updates can be performed frequently for two reasons. First, we only require a few annotated samples for the update. Second, the prototype vectors are compact enough to consume minimal energy during updates. The details of updating the prototype are discussed in detail in Section~\ref{sec:method:hardware}, where we explain Challenge~\circled{3} within the hardware system.

% This two-phase methodology allows us to effectively train, deploy, and update our model in a real-world scenario, while also facilitating the collection of valuable data for further refinement of the model.

% During this phase, we utilize the fine-tuning set as the support set and then select extra support samples from the remaining dataset. The test classes are taken as query set. The episodes undergo multiple passes through the DNN, during which feature vectors are extracted for each sample, and prediction probabilities are computed for all available classes. The final classification is determined by taking the mean of these multiple prediction probabilities.

\subsection{Hardware architecture in \ours}
\label{sec:method:xheep}
\begin{figure}[ht]
    \centering
    \includegraphics[width=1\linewidth]{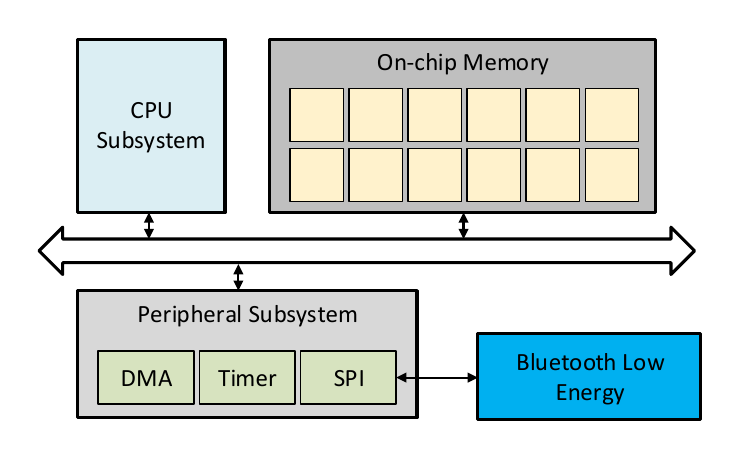}
    \caption{Hardware structure in X-HEEP. In the MetaWear system, the updated prototypes are transmitted to the CPU through the BLE module.}
    \label{fig:method:hardware}
\end{figure}

The system considered in our scenario for wearable systems is based on X-HEEP{~\cite{machetti2024xheep}}, an open-source RISC-V extensible and configurable platform for edge-computing. Shown in Fig.\ref{fig:method:hardware}, X-HEEP comprises three key hardware components: an OpenHW Group CV32E40P RISC-V processing unit (CPU){~\cite{gautschi2017near}}, an on-chip memory, and a peripheral subsystem, all interconnected via a bus. The CPU implements RV32IMFC RISC-V extensions and is designed to deliver high performance while maintaining energy efficiency, a critical requirement for wearable devices due to their limited battery capacity.

The peripheral unit of our system encompasses several subsystems, one of which is the Serial Peripheral Interface (SPI) unit. This unit is tasked with sending and receiving data from an external wireless communication system. In our specific scenario, this unit receives all updated models and prototypes from a BLE unit. It is important to note that accessing the data via the SPI takes significantly longer than accessing the on-chip memory. 

The process of updating the model or prototype, as received from the server side, unfolds as follows: The updated model weights or prototypes start arriving from BLE. The CPU then fetches these values using the SPI protocol and replaces the corresponding parameters in the on-chip memory. 
% This process is carried out between two detection processes, during the idle time while the system waits for the new input signal to fill its buffer. 

In our scenario, the prototype is considerably shorter in length than the entire model's weights. This characteristic allows us to update the prototypes with much less energy consumption, considering that, in the updating process, the bottleneck is the BLE throughput. Note that updates of all the model weights are still possible, but they become unnecessary save for large scenario changes (e.g., a change in monitoring conditions or setup).

The model is updated via a BLE-based method for the X-HEEP architecture. Updates are received by the BLE module and transmitted to the CPU via SPI protocol for updating on-chip memory. The bottleneck in this system is the BLE 4.2’s maximum data transfer rate of 1 Mbps. 

% However, with the proposed method, update times could be reduced by factors of 456x and 418x for epilepsy and AF detection, respectively.
% However, our system is also capable of updating the model's weights in the event of a significant model update. Thanks to our framework, such updates are not required frequently, thereby optimizing energy usage and system efficiency.

%% file: sections/experimental_setup.tex
\section{EXPERIMENTAL SETUP}
\label{sec:setup}
As mentioned in Section ~\ref{sec:intro}, we perform experiments on two different case studies of epileptic seizure detection and AF arrhythmia detection. In the following, we describe the datasets and settings used in these case studies.

\begin{figure*}[ht]
\centerline{\includegraphics[width=0.99\linewidth]{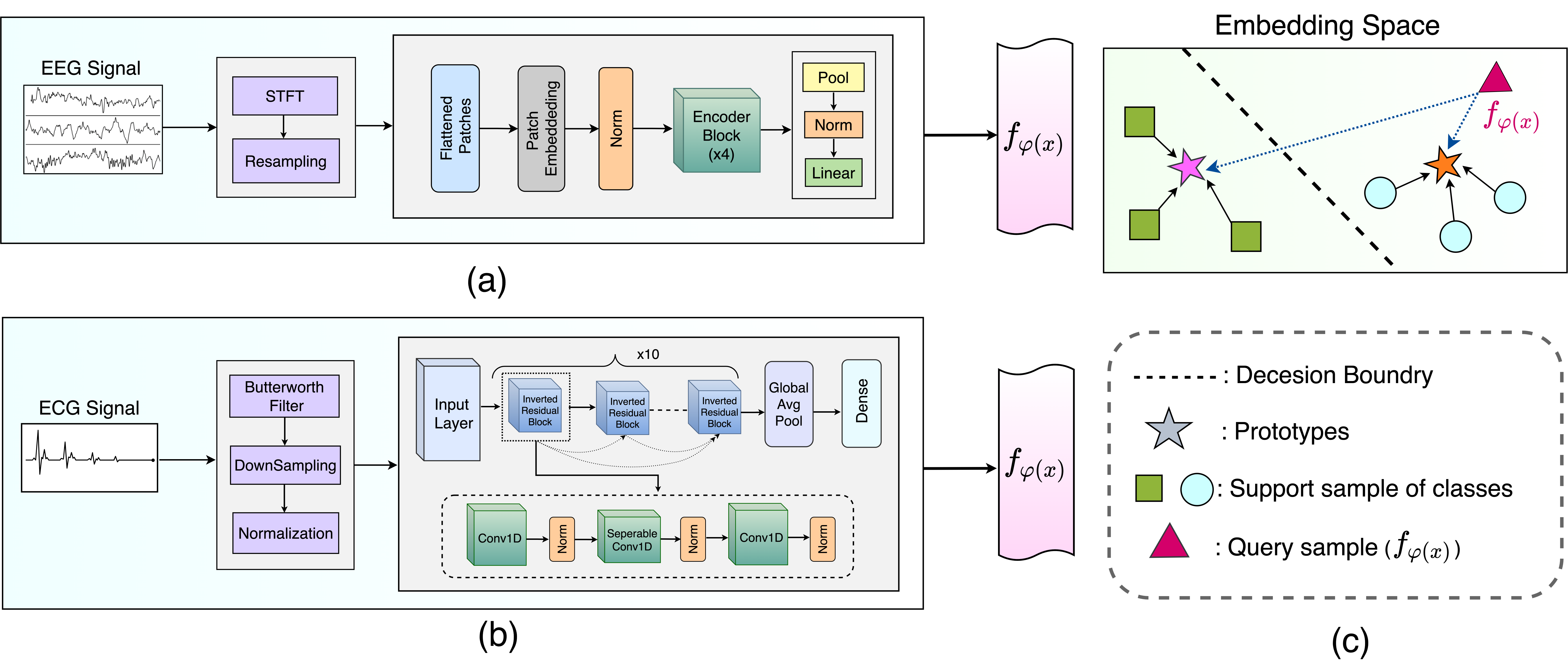}}
\caption{ Overview of model architectures for signal classification in wearable health monitoring :(a)  The EEG signal processing pathway employs a deep learning-based transformer architecture for Epilepsy detection from EEG signals, (b) The AF detection model utilizes a MobileNetV2  neural network to analyze ECG recordings for real-time classification and (c) after computing the feature vector $f_{\varphi(x)}$ for each query sample, these vectors are embedded into a high-dimensional space. In this visual representation, data points are classified by computing their Euclidean distances to class prototype vectors. }
\label{model-architecture}
\end{figure*}

\subsection{EEG data for epileptic seizure detection}
\label{sec:method:eeg}
The training dataset for the base learner in our study comprises the publicly available Temple University Hospital EEG Seizure Corpus (TUSZ){~\cite{obeid2016temple}}-v2.0.0, encompassing data from 675 subjects with a cumulative duration of 1476 hours. Notably, the dataset exhibits imbalances, characterized by predominantly short files (average duration of 10 minutes) and a significant preponderance of normal non-seizure signals. Additionally, the dataset features heterogeneity in sampling frequency and the number of channels. To ensure uniformity, all files are resampled to 256 Hz as suggested in{~\cite{dan2024szcore}}. The spatial arrangement of channels follows the 10-20 system, and we adopt the bipolar montage{~\cite{britton2016electroencephalography}}.

The second database comprises EEG recordings from a limited number of 14 subjects acquired at the Unit of Neurology and Neurophysiology of the University of Siena{~\cite{detti2020siena, detti2020eeg}}. In particular, four subjects lack seizure data and, consequently, they are always used in the test set. The original signals were captured at 512 Hz, and we resampled them to 256 Hz for consistency.

\subsubsection{Data preprocessing and network architecture}
The EEG signals in the base and target datasets are filtered by a band-pass filter on [0.5--60] Hz, as well as a 50 Hz notch filter.  Subsequently, we extract a short-time Fourier transform~(STFT) from each 12-second window of the filtered signal. The STFT employs one-second segmenting, 50 overlapping samples, and a frequency resolution of 2. These parameter choices are extracted by the recommendations in {\cite{ma2023tsd}}.

The model used for this task is a 4-layer VisionTransformer-based model{~\cite{dosovitskiy2020image}}, which is modified for epileptic seizure detection by{~\cite{ma2023tsd}}. The STFT extracted from the EEG input signal is considered as an input image to this 4-layer transformer encoder. Originally, the decoder was implemented as a fully connected layer, reducing the dimensionality to match the number of classes. However, in our approach, using the proposed method, we modify the decoder to employ a fully connected layer, aligning the output dimension with that of the prototypes, which, in this case, is set to 16. Fig.~\ref{model-architecture}a displays the preprocessing steps and model architecture that were utilized for the  Epilepsy detection.

\subsection{ECG data for AF detection}
\label{sec:method:ecg}

%\subsubsection*{Datasets}
The large-scale dataset used for base dataset is Physionet Computing in Cardiology Challenge 2017{~\cite{clifford2017af}}, which consists of 8528 single-lead ECG recordings, divided among AF, normal, noisy, and other classes. We have extracted 7561 signals which have more than 30 seconds of signal. The dataset consists of 5154 normal recordings, 771 AF recordings, 46 noisy recordings, and 2557 records in the other rhythm classes. These recordings have a sampling frequency of 300 Hz. The duration of ECG recordings varies from 9 seconds to 60 seconds, with a median recording length of 30 seconds. 

The second dataset is a private dataset in which data are obtained by a wearable device. This dataset contains 303 records of single-lead ECG signals, with each record having a duration of 40 seconds. The sampling rate of the dataset is 200 Hz. This dataset contains ECG signals for normal heart rhythms and AF. From this dataset, 150 ECG samples were selected for fine-tuning, and the remaining data was reserved for testing.

\subsubsection{Data Augmentation}
Within our base training dataset, a total of 7561 ECG signals with a recording length of 30 seconds are present. However, this dataset demonstrates a pronounced imbalance, with approximately 89\% of the data affiliated with a specific class, leading to a potential bias towards the class with the most data.
% Different strategies can be employed for data augmentation. 
In our study, we choose a combined approach involving random resampling and the Gaussian noise method. Initially, random resampling is used to balance the AF and Other classes, rectifying any imbalances. 
% Following this, the Gaussian noise method is applied to create new data points. Gaussian noise data augmentation is a straightforward and efficient method to expand dataset size and diversity artificially. 
This technique selects samples from the minority class at random and integrates them into the training data. Furthermore, it introduces noise to the synthetic data points, improving variation and smoothing class boundaries, which in turn helps reduce overfitting.
After balancing the number of data in each class, we added amplified noise to only half of the data. 

 \subsubsection{Data preprocessing and network architecture}
Data preprocessing is an essential step in ECG signal analysis, enhancing data quality, reducing noise, and facilitating reliable analysis and modeling. These preprocessing steps help to improve the quality of the ECG signals and reduce inconsistency across different datasets, as in {\cite{kim2022lightweight}}. This method involves the application of a second-order band-pass Butterworth filter to eliminate baseline drift and high-frequency noise from ECG signals. Following this, downsampling at 100 Hz is used to ensure uniformity in sampling rates across different datasets. The signals are then normalized using min-max scaling, resulting in values within the range of 0 to 1.

The CNN structure employed in this study draws inspiration from MobileNetV2 architecture and the architecture in {\cite{kim2022lightweight}}, where a 7-layer instantiation of MobileNetV2 is optimized for AF detection. Based on the memory capacity in our hardware system, we increased the number of layers to 13 layers.

The employed model consists of ten inverted residual blocks connected in sequence. Each inverted residual block includes a depthwise convolutional layer, followed by a projection layer that reduces the channel size back to the original size. The number of output channels for each inverted residual block is designed to be 8, 12 (x2), 16 (x3), and 24 (x4), respectively.
The activation function used in each convolutional layer is the Rectified Linear Unit (ReLU).
The model also includes a separable 1D convolutional layer (separable Conv1D) with channel expansion and a pointwise convolutional layer with six times extended channels. The kernel size for the separable Conv1D layer and the inverted residual blocks is set to 7 and 5, respectively.
The last layer of the model is a dense layer followed by a relu activation function, which performs the classification into the four ECG classes: normal sinus rhythm, atrial fibrillation, atrial premature contraction, and ventricular premature contraction.
Overall, the proposed CNN structure is designed to be lightweight and efficient for ECG classification on low-power wearable devices. Fig.\ref{model-architecture}b displays the preprocessing steps and model architecture that were utilized for the AF detection.

Fig.~\ref{model-architecture}c illustrates the classification process. First, the prototype vectors are computed. Then each query sample is embedded into a high-dimensional space and classified according to its distances from the prototypes. This highlights the model's ability to distinguish between classes.

%% file: sections/result.tex
\section{RESULTS}

\subsection{Primary data collection and training}
\label{sec:exp:phase_1}
In the first phase of a wearable system lifecycle, wearable data must be collected from the patients and must be annotated. These labeled signals form the ``target dataset," in which all the signals are derived from the target wearable system. We assume that the amount of labeled signals is too small for training a model. In this section, we tackle Challenge \circled{1} to meta-train the model only with a minimal amount of annotated signals in the target dataset.

In our work, we use a ``base dataset", which contains a large number of signals with the same type as the target dataset, albeit in a different setting. In particular, for the epileptic seizure detection task, both the base and target datasets have EEG signals, but the base dataset is collected with hospital setting devices. Similarly, in the case of AF detection, the type of signal in the base and target datasets is ECG, with the base dataset gathered in a hospital setting and the target dataset from wearable systems. 

\begin{figure}[t]
    \centering
    \includegraphics[width=0.9\linewidth]{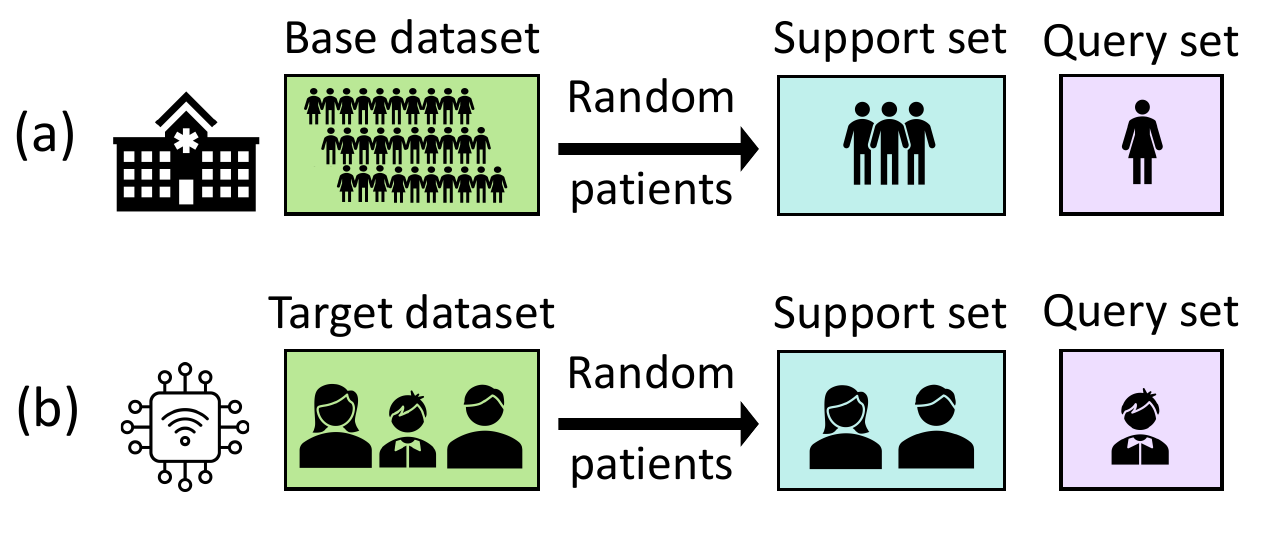}
    \caption{(a) Pretraining and (b) fine-tuning the model on the meta-train step in \ours. This figure also shows the split of data into support set and query set in each episode.}
    \label{fig:method:phase1_training}
\end{figure}

Fig.~\ref{fig:method:phase1_training}{a} shows the base dataset. The base dataset is an extensive dataset, and it contains signals acquired from hospital setting devices. From a large pool of patients, for each episode, we randomly select non-overlapping patients, for the support set and query set, respectively. We randomly choose samples from these patients to form a support set and a query set to pretrain the model in the base dataset.

\begin{figure}[t]
\centering
\includegraphics[width=0.9\linewidth]{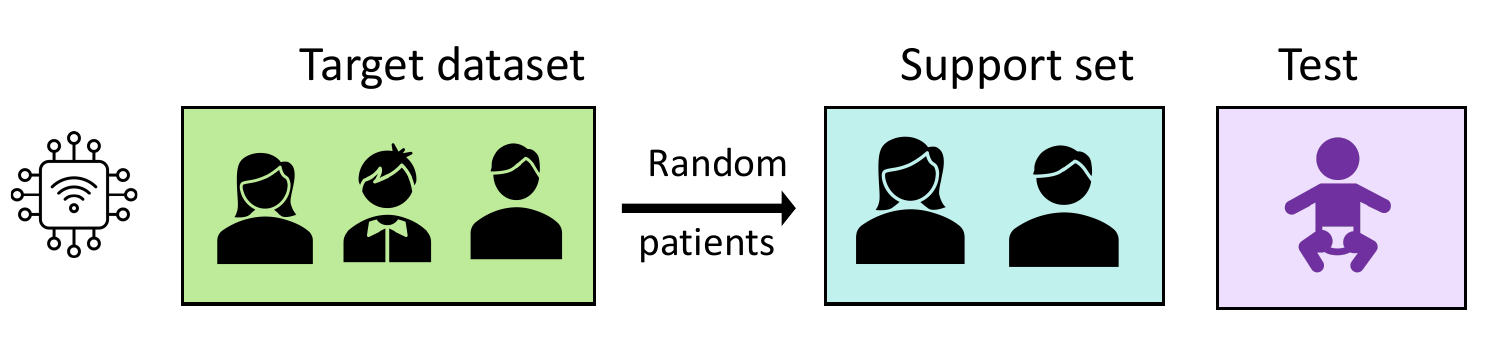}
\caption{Inference process on the wearable systems. Although the amount of labeled data in the target dataset is limited, thanks to \ours{} we can use only a few shots for training the model and address Challenge \protect\circled{1}.}
\label{fig:method:phase1}
\end{figure}

Fig.~\ref{fig:method:phase1_training}{b} shows the target dataset, where signals are collected from wearable devices and annotated by clinicians. The number of patients in this target dataset is significantly smaller. Here, we fine-tune the model based on this target dataset. Again, we apply the random sampling of patients to form the support set and the query set in each episode. Through fine-tuning, the model becomes adapted to wearable signals.

In the inference meta-test step, as illustrated in Fig.\ref{fig:method:phase1}, random patients and random samples are selected from the target dataset to create the support set and the test sample is derived from a non-overlapping patient.

\subsubsection{Quantitative results}
\begin{figure}[t]
\centering
\begin{subfigure}{.24\textwidth}
  \centering
  \includegraphics[width=1\linewidth]{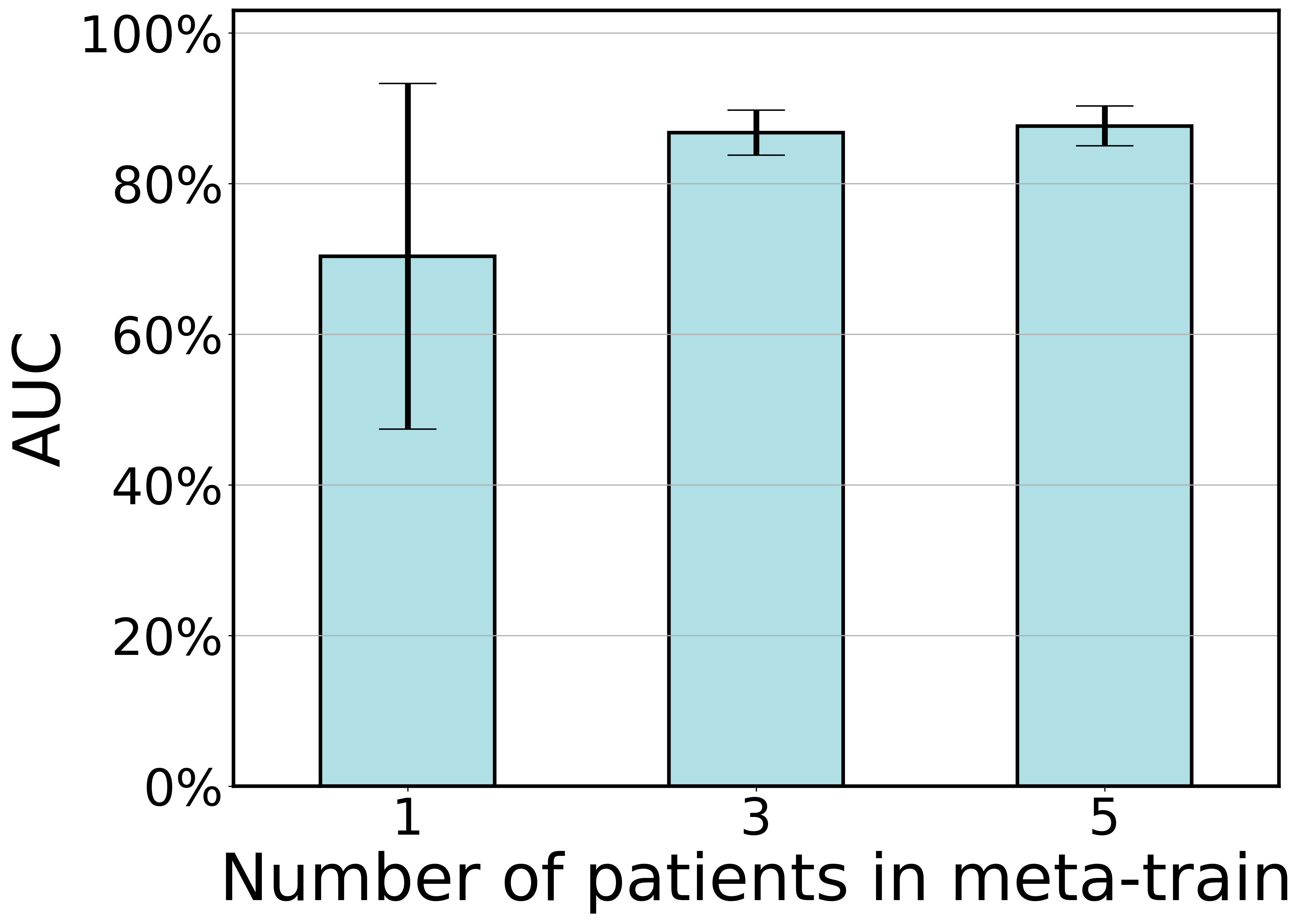}
   \subcaption{Epileptic seizure detection}
  \label{fig:phase_1_Epilepsy}
\end{subfigure}%
\begin{subfigure}{.24\textwidth}
  \centering
  \includegraphics[width=1\linewidth]{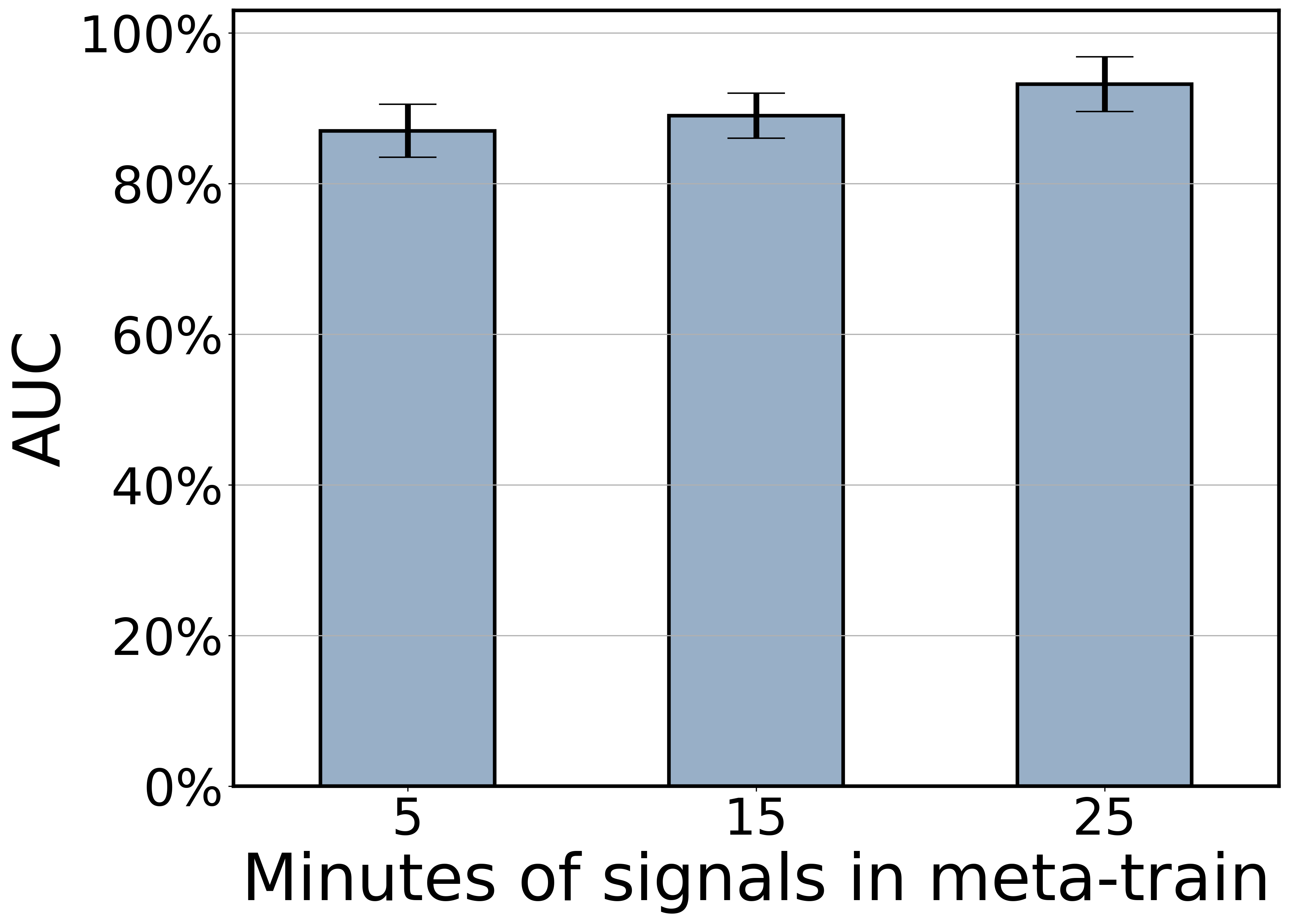}
   \subcaption{AF detection}
  \label{fig:phase_1_af}
\end{subfigure}
\caption{The model's performance based on the amount of labeled data in the target dataset. This figure illustrates the efficacy of \ours{} even with very limited labeled data in the target dataset, addressing Challenge \protect\circled{1}.}
\label{fig:phase1}
\end{figure}

Fig.~\ref{fig:phase1} shows the results of our experiments based on the amount of labeled data in the target dataset. 
For epilepsy, we have selected annotated groups from the target dataset with one, three, and five patients, and similarly, for AF, we have selected annotated groups with varying ECG signal durations of 5, 15, and 25 minutes to investigate the effect of the number of initial annotated signals on the performance of the model. As shown in Fig.~\ref{fig:phase1},  the results have improved consistently as the number of samples within each group has gradually increased.

During pretraining on the base dataset, in each episode, we set a query set size of 15 from two classes in the AF large-scale dataset and two classes in the epilepsy base dataset. At test time, we similarly sample episodes from target classes and average the results over 10 iterations to get trustworthy measures of the model performance.
% We use the Adam optimizer with a learning rate of 0.0001 for each of the learning models.
During fine-tuning on the target datasets, in each epoch, there are 5 episodes for AF and 100 episodes for epilepsy and then selected $k=5$ samples for the support set. Also, 20\% of the data are selected as validation. In case where the validation loss is not decreased for five epochs, the fine-tuning is stopped.

In the meta-test phase, we perform multiple iterations to evaluate our model's performance under varying conditions. For each iteration, the model is applied to the test data, resulting in a set of the Area Under the ROC Curve (AUC). We then report the mean AUC value. 
% In addition, we computed the standard deviation (STD) to measure the degree of dispersion or variability in these AUC scores. This iterative process, along with reporting mean AUC and STD, ensures a thorough and statistically accurate assessment of our model's performance.

As shown in Fig.~\ref{fig:phase_1_Epilepsy}, by increasing the number of patients in each group, the AUC value increases by 16\%. With fine-tuning on only five patients, our model obtained an AUC value of 87.7\%.
Similarly, in Fig.~\ref{fig:phase_1_af}, by increasing the time of the ECG signals in each group, we have an increase in the AUC value by 2\% and then 4\% for 15 and 25 minutes of data in fine-tuning, respectively. Finally, with fine-tuning on 25 minutes of signals, our model's AUC value is 93\%.

\begin{figure}[t]
\centering
\begin{subfigure}{.24\textwidth}
  \centering
  \includegraphics[width=1\linewidth, height=4cm]{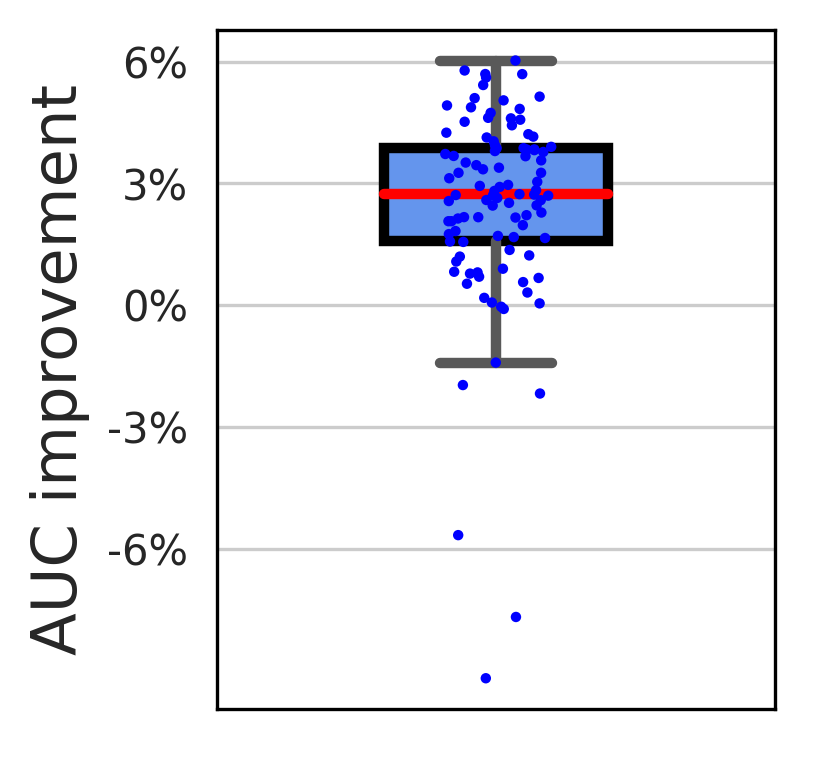}
   \subcaption{Epileptic seizure detection}
  \label{fig:bp_Epilepsy}
\end{subfigure}%
\begin{subfigure}{.24\textwidth}
  \centering
  \includegraphics[width=1\linewidth, height=4cm]{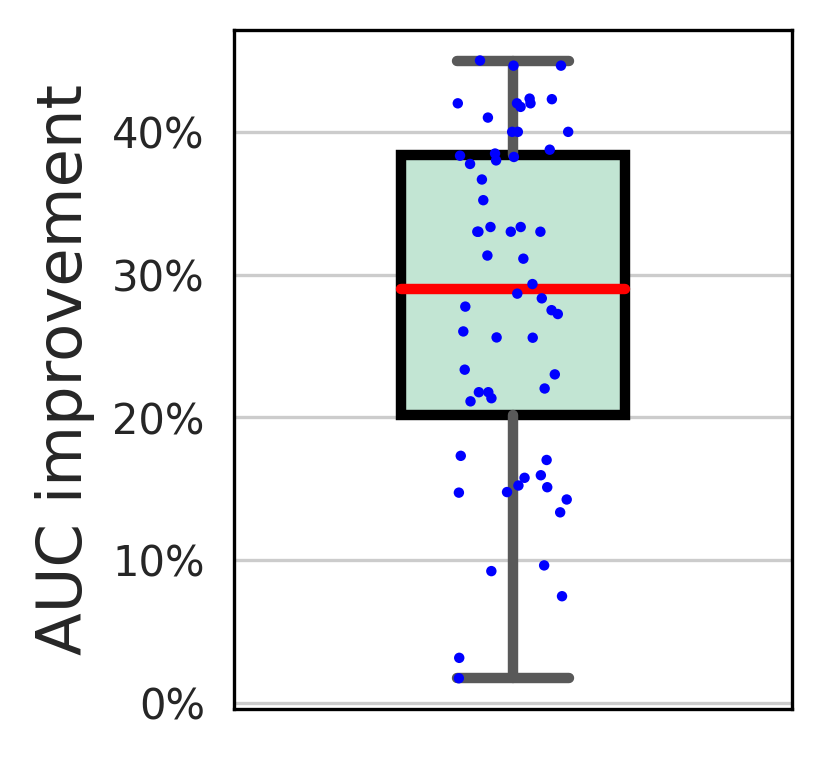}
   \subcaption{AF detection}
  \label{fig:bp_AF}
\end{subfigure}
\caption{AUC improvement for seizure detection and AF detection comparing Source-only and MetaWears models.}
\label{fig:box_plot}
\end{figure}

% \begin{figure}[t]
%     \centering
%     \includegraphics[width = 0.9\linewidth]{figures/sota_box_plot.png}
%     \caption{AUC improvement for seizure detection and AF detection comparing Source-only and MetaWears models. }
%     \label{fig:box_plot}
% \end{figure}

\subsubsection{Comparison with previous studies}

As discussed in the previous section, Challenge \circled{1} is related to the limited data available in the target dataset for training an ML model. In previous works, this challenge is often tackled by training a model on a larger dataset (base dataset) and directly evaluating it on the target dataset. This approach has been commonly used in various medical research studies that involve various input data, including medical images~{\cite{radhakrishnan2023cross, shad2021designing, li2023towards, faes2019automated}}, biomedical signals~{\cite{lai2023practical,turbe2023evaluation, siontis2024detection,hannun2019cardiologist, amirshahi2022m2d2, saab2020weak}}, electronic health records~{\cite{rasmy2021med, kraljevic2024foresight,barish2021external,rajkomar2018scalable}}, and combinations of these~{\cite{yin2021accurate,diaz2022predicting,pirruccello2024deep,wang2022high}}.

This section compares the performance of \ours{} with this conventional approach. To ensure a fair comparison, we conducted multiple experiments based on different patient combinations within the target dataset. We then obtained the corresponding results for both \ours{} and the conventional approach. The AUC improvement using \ours{} for each target dataset was then calculated as the subtraction between \ours{} result with the convention approach one. Fig.~\ref{fig:box_plot} presents the distribution of AUC improvement for various patient combinations in the target dataset for both seizure and AF detection tasks.

As shown in the figure, the median AUC improvement for \ours{} compared to the conventional approach is 2.7\% and 29.0\% for seizure and AF detection, respectively. To further analyze this, a one-sample t-test was performed to determine if the AUC improvement is significantly greater than zero. The null hypothesis assumed that zero belongs to the distribution. The obtained p-value was less than 0.05, leading to the rejection of the null hypothesis. This evidence confirms that the AUC improvement is statistically greater than zero, indicating that the \ours{} model outperforms the conventional approach in the studied detection tasks.

The results reveal a greater performance difference between \ours{} and the conventional approach for AF detection compared to seizure detection. In particular, the AF detection task does not exhibit instances of extreme under-performance or outliers. This observation can be attributed to the larger discrepancy in signal features between the target and base datasets for AF detection. This is because the AF task utilizes data from a wearable device in the target dataset, whereas seizure detection utilizes data from a hospital setting, different from the base hospital data. This issue is further discussed in Section~\ref{sec:limitation}.

% \begin{figure}[t]
% \centering
% \begin{subfigure}{.24\textwidth}
%   \centering
%   \includegraphics[width=1\linewidth]{figures/seizure_box_plot.png}
%    \subcaption{Epileptic seizure detection}
%   \label{fig:Box_plot_Epilepsy}
% \end{subfigure}%
% \begin{subfigure}{.24\textwidth}
%   \centering
%   \includegraphics[width=1\linewidth]{figures/af_box_plot.png}
%    \subcaption{AF detection}
%   \label{fig:Box_plot_AF}
% \end{subfigure}
% \caption{AUC improvement for (a) Epilepsy and (b) AF comparing Source-only and MetaWears models. }
% \label{fig:results:source_only}
% \end{figure}

\begin{figure}[t]
    \centering
    \includegraphics[width=1\linewidth]{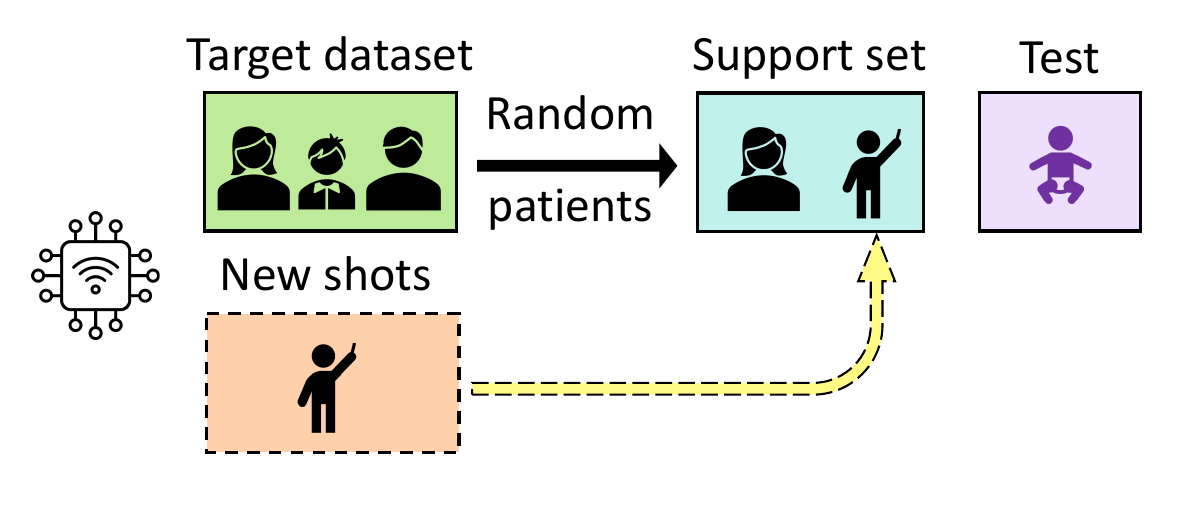}
    \caption{Updating workflow for enhanced wearable system performance using only a few new shots to tackle Challenge \protect\circled{2}.}
    \label{fig:method:phase2}
\end{figure}

 \begin{figure}[ht]
\centering
\begin{subfigure}{\columnwidth}
  \centering
  \includegraphics[width=0.99\linewidth]{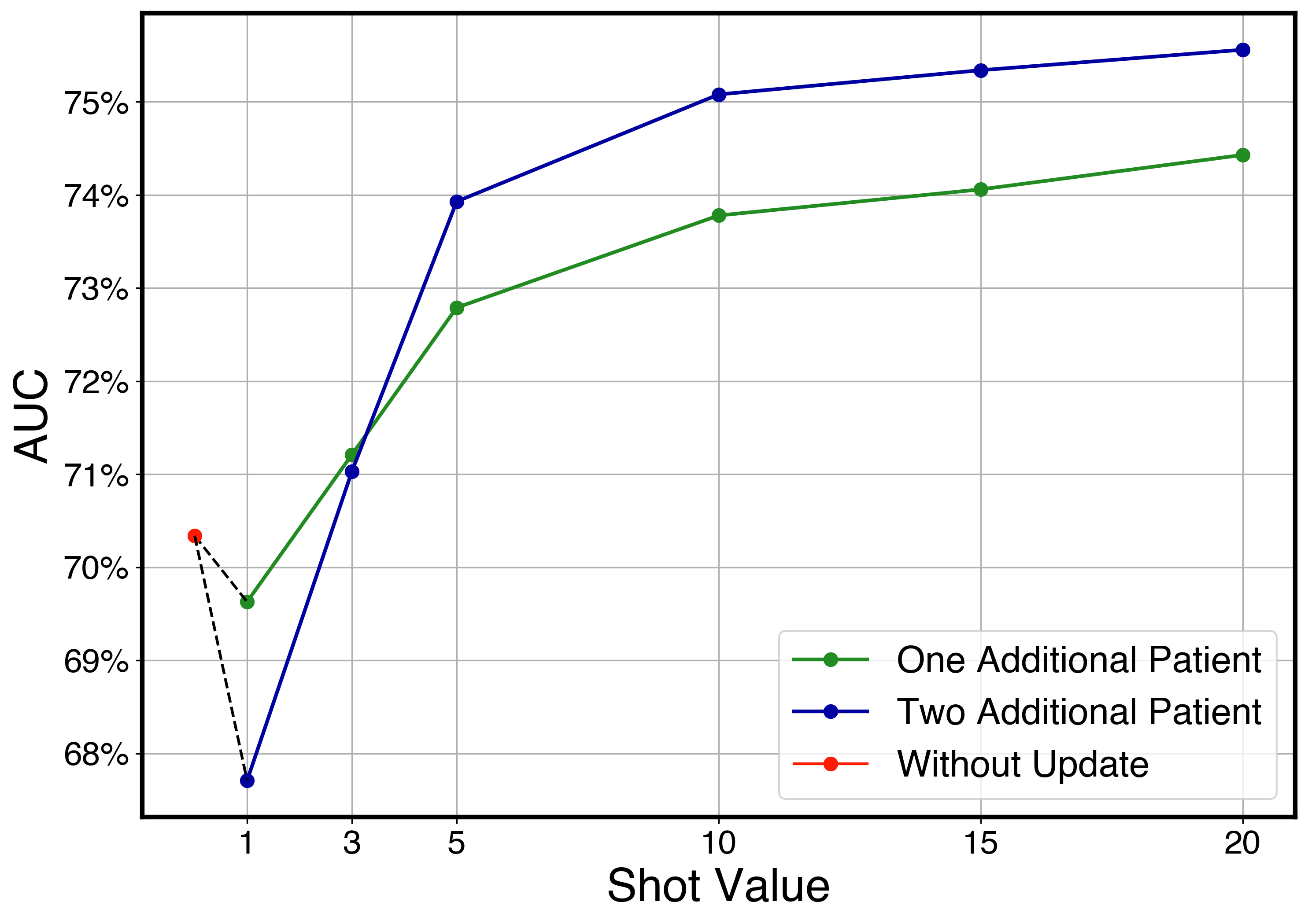}
   \subcaption{Epileptic seizure detection}
  \label{fig:phase_2_Epilepsy}
\end{subfigure}

\begin{subfigure}{\columnwidth}
  \centering
  \includegraphics[width=0.99\linewidth]{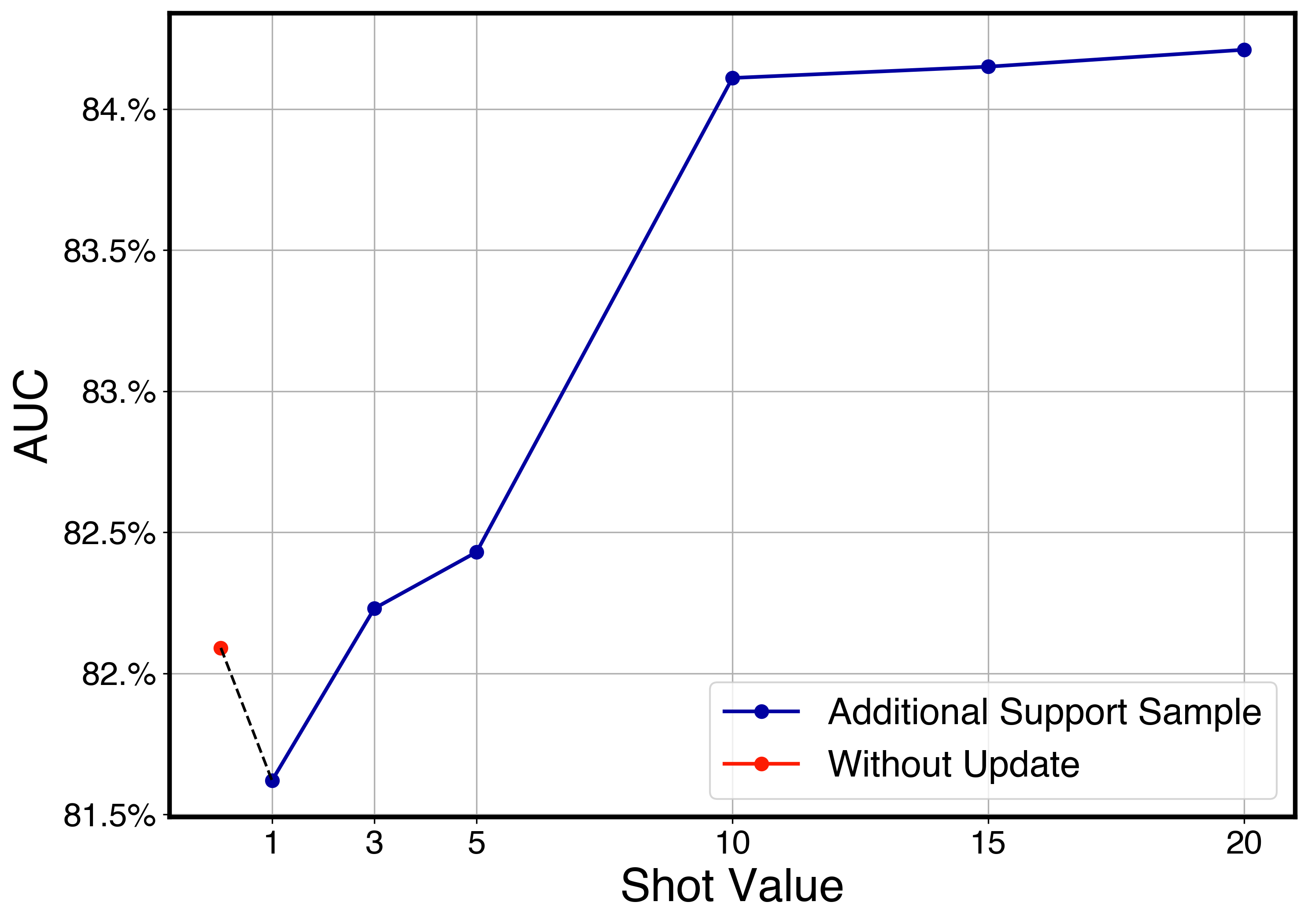}
   \subcaption{AF detection}
  \label{fig:phase_2_AF}
\end{subfigure}
\caption{Improving model performance using new few shots to address Challenge \protect\circled{2}. In both cases, AUC is increased by updating prototypes when $k\geq3$ new shots are received.}
\label{fig:results:phase2}
\end{figure}

\subsection{Updating the model using few new shots}

In the second phase of wearable system lifecycles, the wearable devices equipped with our model are distributed to new patients for detection purposes.  The wearable systems begin collecting signals from the individuals. The system is designed to detect target abnormalities and transmit this data to a server for further processing. As data is accumulated from the wearable systems, clinicians are able to annotate the signals. In this way, a new labeled sample is available for updating the model.  In this section, we discuss the reduction of the number of new labeled data for updating the model using \ours{} to address Challenge~\circled{2}.

As shown in Fig.~\ref{fig:method:phase2}, the new shot is included in the support set that was also created with samples from the target dataset. Therefore, the support set is updated, and the new few shots play a crucial role in the inference step.

To evaluate the effectiveness of our model update scenario for the detection of AF and epilepsy, we conducted experiments to evaluate the results of the phase where prototypes are calculated based on the newly received annotated shots, typically on the server side. For AF, we performed fine-tuning on the group that included the smallest number of patients, which is equal to 30 ECG records and takes 5 minutes, and then we only took a few shots from other groups of patients that are not part of the test data in the meta-test phase. 
For epilepsy, fine-tuning was performed in one patient and then new shots were taken from one and two additional patients. These experiments involved varying the shot values of $k=1, 3, 5, 10, 15,$ and $20$.
% , and the hyperparameter tuning is similar to Section~\ref{sec:exp:phase_1}.

As shown in Fig.~\ref{fig:phase_2_Epilepsy}, for epileptic seizure detection, the baseline AUC is 70.3\%, where no additional shots are applied. When $k=10$ shots are received from a single additional patient in $\mathcal{D}^{\text{new}}$, there is an improvement in the AUC by 3.4\% compared to the baseline. This improvement is even more pronounced when $k=20$ shots are received from the same patient, resulting in 74.4\% of AUC. 

The figure also presents the results when two additional patients shared their shots. With $k=20$ new shots from these two patients, the model can achieve 75.6\% of AUC, which is 5.3\% more than the baseline, only by a total of 16 minutes of signals.
%To classify a new sample of epilepsy, the first test we performed was that we did fine-tuning with only limited samples from one patient by applying new shots that can be received in the wearable device. The model is updated immediately, and the result shows that for a new sample, AUC is 68\%, and with the addition of more shots, the result increases to about 74\%.
% In another test, we want to classify the signals of two additional patients while still having one patient in the fine-tuning phase. At this stage, the new sample is again fine-tuned by adding to the initial data.
% In another test, we want to classify the signals of two new patients while still having one patient in the fine-tuning phase. At this stage, the new sample is again fine-tuned by adding to the initial data, and the model will be updated.
% The results are shown in Figure~\ref{fig:phase_2_Epilepsy}.

% \begin{figure}[t]
% \centering
% \begin{subfigure}{.24\textwidth}
%   \centering
%   \includegraphics[width=1\linewidth]{figures/epilepsy_phase2.png}
%    \subcaption{Epileptic seizure detection}
%   \label{fig:phase_2_Epilepsy}
% \end{subfigure}%
% \begin{subfigure}{.24\textwidth}
%   \centering
%   \includegraphics[width=1\linewidth]{figures/af_phase2.png}
%    \subcaption{AF detection}
%   \label{fig:phase_2_AF}
% \end{subfigure}
% \caption{Improving model performance using new few shots to address Challenge \protect\circled{2}. In both cases, AUC is increased by updating prototypes when $k\geq3$ new shots are received.}
% \label{fig:results:phase2}
% \end{figure}

An observation in Fig.~\ref{fig:results:phase2} in both cases of epileptic seizure detection and AF detection is the slight performance degradation compared to the baseline when only a single shot is received from one or two additional patients. The reason for the performance degradation can be the low robustness of randomly selecting a single shot as opposed to multiple shots. This phenomenon is explained in more detail in Section~\ref{sec:discuss:visual}.

\begin{table*}
    \centering
    \begin{small}
    \caption{Comparison of low-latency and low-power modes of operation for the epilepsy and AF detection applications.}
    \label{tab:power-mode-comparison}
    \begin{tabular}{c|c|c|c|c|c|c|c|c}
    \toprule
    \toprule
       % \rowcolor{table_color_1} 
        & & Frequency & Active power & Idle power & Voltage & Window & Exec. time & Battery life\\
       % \rowcolor{table_color_1} 
        & & [MHz] & [mW] & [mW] & [V] & [seconds] & [seconds] & [hours]\\
        \hline
        \hline
         \multirow{2}{*}{low-latency} & Epilepsy & 450 & 50 & 18.7 & 1.2 & 12 & 1.9 & 24.3\\
         & AF & 450 & 50 & 18.7 & 1.2 & 10 & 0.76 & 27.3\\
         \hline
         \multirow{2}{*}{low-power} & Epilepsy & 75 & 3.7 & - & 0.8 & 12 & 12 & 103.8 \\
         & AF & 34.5 & 1.9 & - & 0.8 & 10 & 10 & 202.1 \\
    \bottomrule
    \bottomrule
    \end{tabular}
    \end{small}
\end{table*}
 % The results of this phase show that the combination of low data requirements, real-time model updates, and efficient prototype management not only enhances system performance without the need for extensive data collection efforts but also leads to remarkable improvements in results, making our method a promising solution for dynamic healthcare scenarios.

\subsection{Battery life with \ours}
Addressing Challenge~\circled{3} that considers the energy efficiency of the updates, we use X-HEEP as the energy-efficient and low-power system for implementing the models{~\cite{machetti2024xheep}}. 

The X-HEEP hardware supports both low-latency and low-power modes, detailed in Table \ref{tab:power-mode-comparison}. In low-latency mode, the system uses a 1.2V supply and 450 MHz frequency, allowing rapid execution and subsequent idling. The transformer model used for epileptic seizure detection processes each 12-second window as input. Remarkably, we observe that the processing time for each window is 1.9 seconds when X-HEEP operates at the low-latency mode. Also, in AF detection, we process a 10-second window in 0.76 seconds at the low-latency mode. The estimated power consumption is 50~mW during operations. Using a battery with a capacity of 480 mAh, we can execute the seizure detection and AF detection tasks on X-HEEP for 24.3 and 27.3 hours, respectively.
Therefore, thanks to the optimized implementation and quantization efforts undertaken for these models, they can operate in real-time on the X-HEEP system. 
 
 Conversely, the low-power mode focuses on reducing power usage by adjusting frequencies to just meet real-time demands, eliminating idle periods. Here, epilepsy detection operates at 75 MHz with 3.7 mW power, and AF detection at 34.4 MHz with 1.9 mW power. A battery with the same capacity of 480 mAh can last in the low-power mode as long as 103.8 and 202.1 hours for seizure detection and AF detection, respectively.

For epilepsy detection, updating the prototypes via the Bluetooth Low Energy (BLE) module takes 239 milliseconds, which is 12\% of the total execution time. For AF detection, updates take 1.7 seconds, amounting to 225\% of the execution time, significantly impacting performance. 

A common approach to updating models on wearable devices involves sending the entire updated model back to users' devices from the server. This method is prevalent in federated learning scenarios, as seen in several works {\cite{dou2021federated, dayan2021federated,bercea2022federated, bo2023relay, baghersalimi2021personalized,wu2022communication}}.

In contrast, our proposed \ours{} method enables us to update the wearable system by only updating the prototypes instead of updating the entire model's weights. This benefit reduces the time and energy consumption for updating the framework by 456x and 418x for epileptic seizure detection and AF detection, respectively. In case of frequent updating of the model, the battery life is decreased significantly in the conventional approach. 
%Therefore, Challenge~\circled{3} that considers the energy efficiency of the updates is addressed by \ours. 

% \subsubsection{\ours{} as a hardware-efficient solution}\label{sec:method:hardware}

% Fig.~\ref{fig:phase_2_AF} shows the results for AF detection. The baseline model, which has been fine-tuned with a minimum time of 5 minutes of data, achieves 82.1\% of AUC. When the prototype is updated with only an additional 3 minutes of data ($k=10$), there is an increase of 2\% in the result.

\subsubsection{Optimizing memory usage in \ours{}-based wearable systems}\label{sec:method:hardware}

Table~\ref{tab:hardware} provides the details of the models used in this study. In the case of epilepsy, the transformer model is deployed on X-HEEP using 16-bit fixed-point parameters. Within the transformer, the linear operations are executed with fixed-point operations, while the non-linear operations, namely Softmax, Normalization, and GELU activation, are operated by a 32-bit floating-point unit. While quantization of the parameters and the linear operations to 16-bit enables us to save data size and increase performance, it has a negligible effect on the results, with only a 0.1\% drop in AUC in the case of epileptic seizure detection using the proposed method.
In particular, the quantized implementation fits entirely in the on-chip memory, which leads to higher energy efficiency because no data movements from the external memory are needed. In addition, quantization to 16-bit integer numbers allows for higher performance on the CPU with respect to floating-point operations.
For the AF detection task, the MobileNetV2 model operates with 32-bit floating-points both for the parameters and operations. Based on these parameters and operation types, we show in Table~\ref{tab:hardware} the sizes of the models, input, intermediate output, and prototypes for these two case studies.

\begin{table}
    \centering
    \begin{small}
    \caption{Hardware implementation detail of the models}
    \vskip -0.1in
    \label{tab:hardware}
    \renewcommand{\arraystretch}{1.5}
    \begin{tabular}{|c|c|c|}
        \hline
         & \textbf{Epilepsy} & \textbf{AF}\\
        \hline
      \multirow{2}{*}{Parameters type} & Fixed-point & Floating-point\\ 
            & 16 bit & 32 bit \\
            \hline
            \multirow{2}{*}{Operations type} & \multirow{2}{*}{Hybrid} & Floating-point \\
            & & 32 bit\\
             \hline
        Model size & 29.2 KB & 209 KB \\
         \hline
        Input size & 117 KB & 3.9 KB\\
         \hline
        Intermediate output size & 93 KB & 146.5 KB \\
         \hline
        Prototypes size & 64 B & 512 B\\
        \hline
    \end{tabular}\\[2ex]
\end{small}
\vskip -0.1in
\end{table}

The synthesized system features 384~KB of on-chip memory divided into 12 memory banks, storing the input signal, the model's weight, the prototypes, and the intermediate output. The intermediate output is generated during the inference process, where the model saves the output of each layer for further processing. The part dedicated to the intermediate output can be overwritten during the computation. This memory optimization strategy is particularly beneficial in wearable systems where memory resources are limited.

%% file: sections/discussion.tex
\section{DISCUSSION}

In this work, we have discussed several significant obstacles that are currently at the cutting edge of personalized healthcare and wearable technology. In Challenge~\circled{1},
% , achieving cross-dataset generalization is crucial for model adaptability and versatility. This aspect was explored through the approaches detailed in {~\cite{tang2023multicenter,wang2022high,saab2020weak}}.  Also, 
% we highlight the issue of insufficient data collected from wearable devices, which often reduces the effectiveness of machine learning models. Further research has been conducted to advance meta-learning strategies that address these challenges in various health monitoring applications. Liu et al.~{~\cite{liu2021metaphys}} introduces MetaPhys, a meta-learning framework that optimizes remote physiological measurements, capable of high-speed inference, which shows promise for edge or wearable devices. Zhu et al.~{~\cite{zhu2022enhancing}} discusses the utilization of model-agnostic meta-learning (MAML) to improve wearable platforms for managing type 1 diabetes through real-time glucose monitoring. 
we highlight the issue of insufficient data collected from wearable devices, which often reduces the effectiveness of machine learning models. Further research has been conducted to advance meta-learning strategies that address these challenges in various health monitoring applications. One study introduces MetaPhys, a meta-learning framework that optimizes remote physiological measurements and is capable of high-speed inference, showing promise for edge or wearable devices~{~\cite{liu2021metaphys}}. Furthermore, a study explores the use of model-agnostic meta-learning (MAML){~\cite{finn2017model}} to improve wearable platforms for real-time glucose monitoring, specifically in the management of type 1 diabetes~{~\cite{zhu2022enhancing}}.The study in ~{~\cite{qiu2020meta}} presents a meta-learning neural network framework for genomic survival analysis, achieving effective predictive modeling with limited data and highlighting key cancer-related genes and pathways. Another work{~\cite{gong2021adapting}} introduces the model named MetaSense, which effectively generalizes across multiple datasets using limited mobile sensing data.
While these papers tackled the challenge of limited data and cross-dataset through meta-learning, they have not taken into account the scenarios where the training set has no data from the target subject. In the context of wearable devices, acquiring annotated data from each user for model training is not only costly but also time-intensive due to the infrequency of abnormal events. In \ours, we have clearly separated the meta-train and meta-test based on individual patients to have the real-world applicability of our proposed algorithm. Furthermore, while the mentioned papers provide foundational insights, they have not considered hardware implementation in their methods, nor have they addressed the ongoing updates to the framework over time, which we identify as Challenge~\circled{2}.

In Challenge~\circled{2}, we introduced \ours{} to update models efficiently with minimal wearable device data. Unlike the other update processes performed in the previous study, such as integrating multi-resolution kernels~{~\cite{lungu2020siamese}} or optimizing memory management through gradient descent~{\cite{karunaratne2021robust}}, \ours{} simplifies the update process. Traditional approaches, like the hardware-software integration for energy efficiency~{\cite{jebali2024powering}} or using the modified hardware-aware meta-learning process~{~\cite{murthy2022learn}}, often increase computational demands and energy consumption. Some recent researches focus on self-powered flexible devices using energy harvesting and low-power designs optimized for healthcare applications like real-time health monitoring~{\cite{ambrogio2018equivalent,yao2020fully,wang2018fully}}. The main issue with these methods is their requirement for a stable supply voltage, which is incompatible with the fluctuating output of miniature energy harvesters. In contrast, \ours{} minimizes both, reducing the execution time and resource use critical in time-sensitive and energy-constrained wearable environments. 
Furthermore, the need for a lightweight and energy-efficient model suitable for deployment in wearable devices was important, a challenge that we defined as Challenge~\circled{3}, which is inefficiency in frequent updates of deep learning models on wearable devices.
The study in  {~\cite{gadaleta2023prediction,kim2022lightweight,huang2023epilepsynet}} tackled the model deployment on wearable devices by customizing solutions specifically for the resource constraints of these platforms.
Also other previous works, federated learning has been employed to update a global model via the Federated Averaging (FedAvg) algorithm, which involves centralized aggregation of local data adjustments. This process continues with iterative local and global optimizations, including the exchange of model parameters between the central server and local nodes{~\cite{dou2021federated}}. In addition, another study introduces Relay Learning, in which models are updated and passed from one clinical site to another sequentially. This process is facilitated by a central server and enhances data privacy by maintaining physical disconnection during model transmission{~\cite{bo2023relay}}.

While previous studies have offered energy-efficient real-time solutions for detecting biomedical abnormalities, our \ours{} proposal takes a more comprehensive approach. Not only does it consider the energy efficiency of the wearable system, it also addresses the entire lifecycle of such systems. We have addressed the challenges of limited data in wearable training sets and updated the model with only a few data points. Moreover, our method has been extended to cover two distinct domains, heart and brain signal abnormality detection.

\subsection{Visual representation of samples}
\label{sec:discuss:visual}

\begin{figure}[t]
\centering
\begin{subfigure}{\columnwidth}
  \centering
  \includegraphics[width=0.9\linewidth]{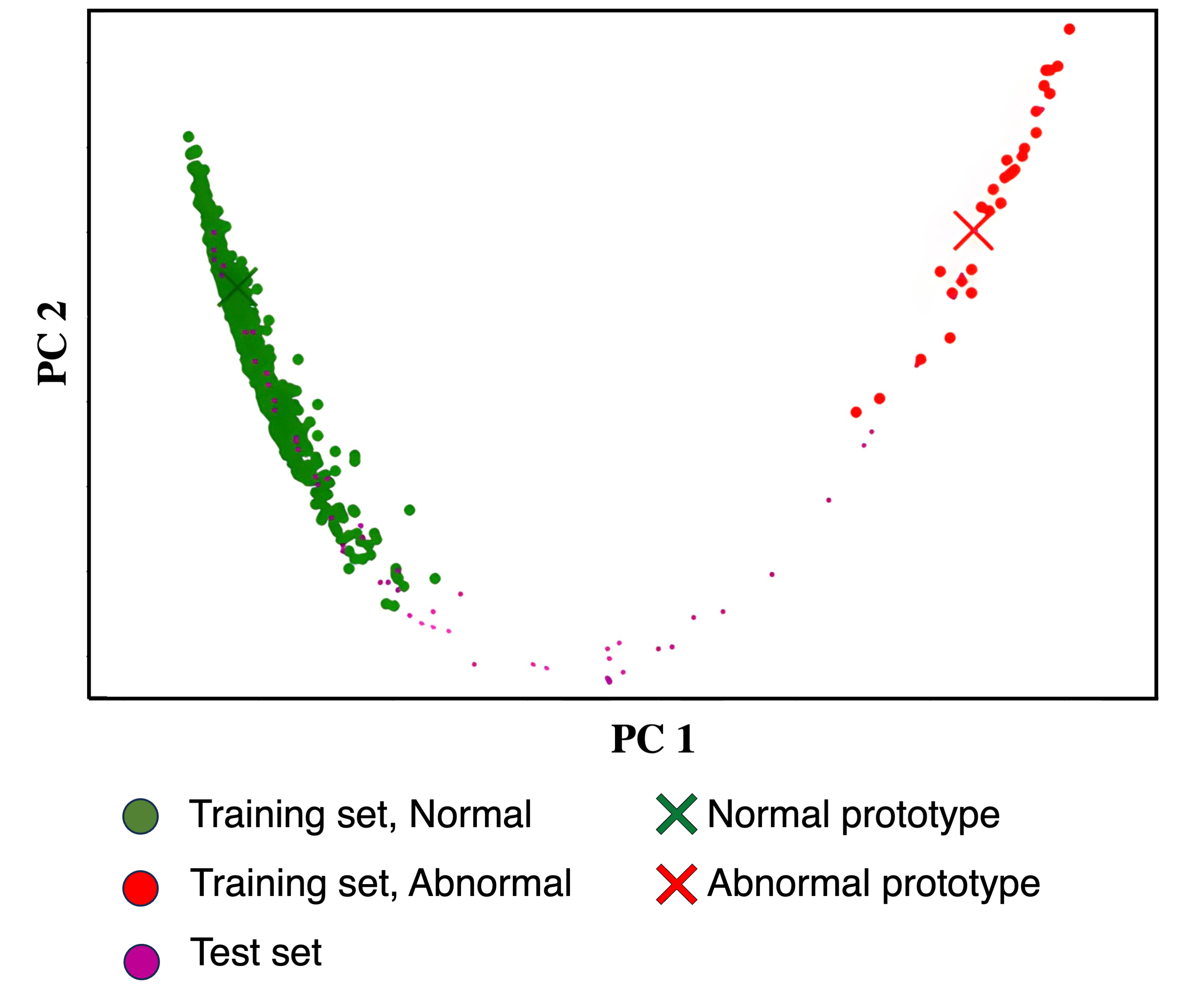}
   \subcaption{Epileptic seizure detection}
  \label{fig:pca_phase_1_Epilepsy}
\end{subfigure}
\begin{subfigure}{\columnwidth}
  \centering
  \vspace{1em}
  \includegraphics[width=0.9\linewidth]{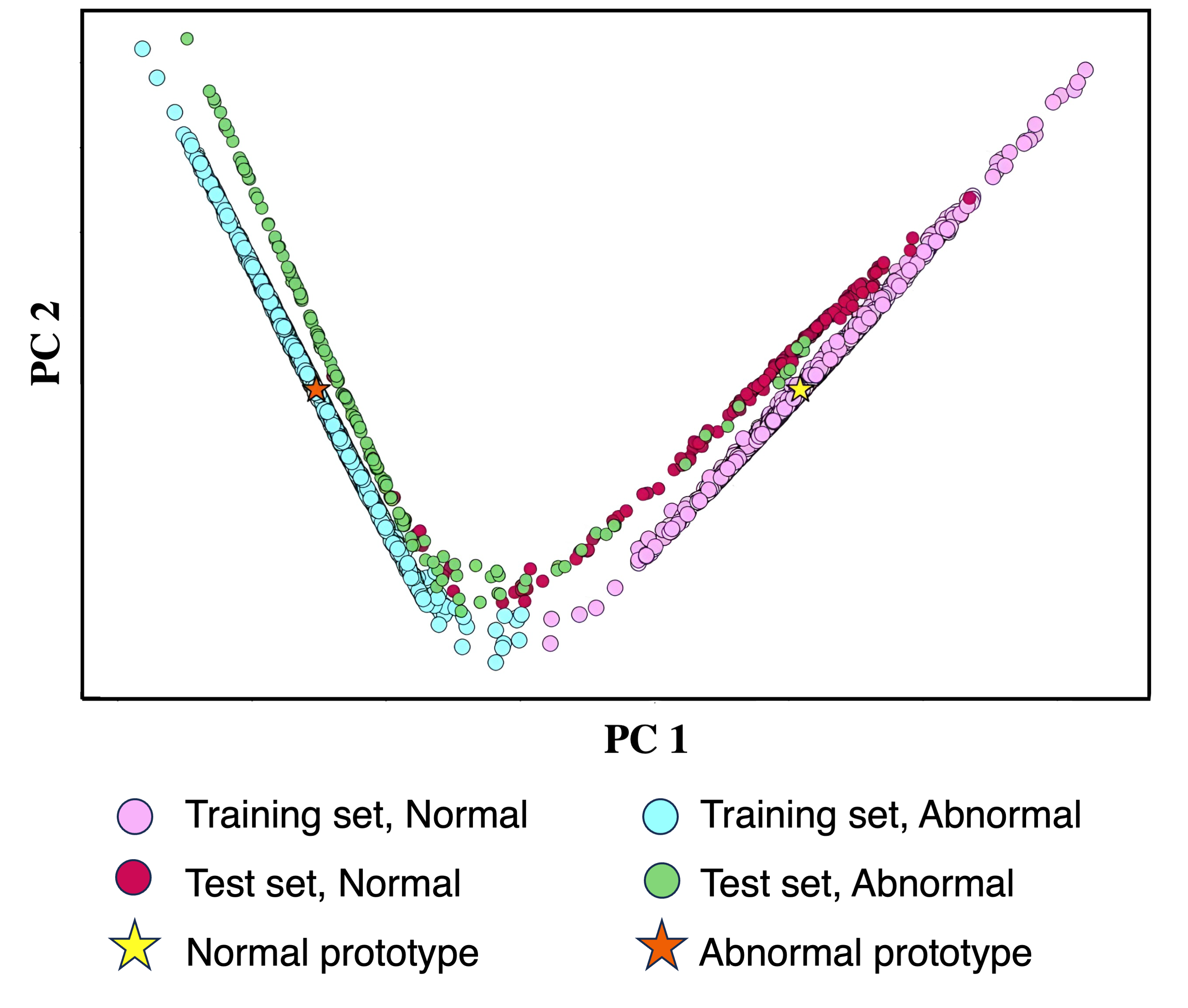}
   \subcaption{AF detection}
  \label{fig:pca_phase_1_AF}
\end{subfigure}
\caption{PCA analysis of support and query data after fine-tuning in response to Challenge \protect\circled{1} for (a) Epilepsy and (b) AF, demonstrating enhanced separability and clustering of the datasets.}
\label{fig:pca_phase1}
\end{figure}

\begin{figure}[t]
    \centering
    \includegraphics[width=0.99\linewidth]{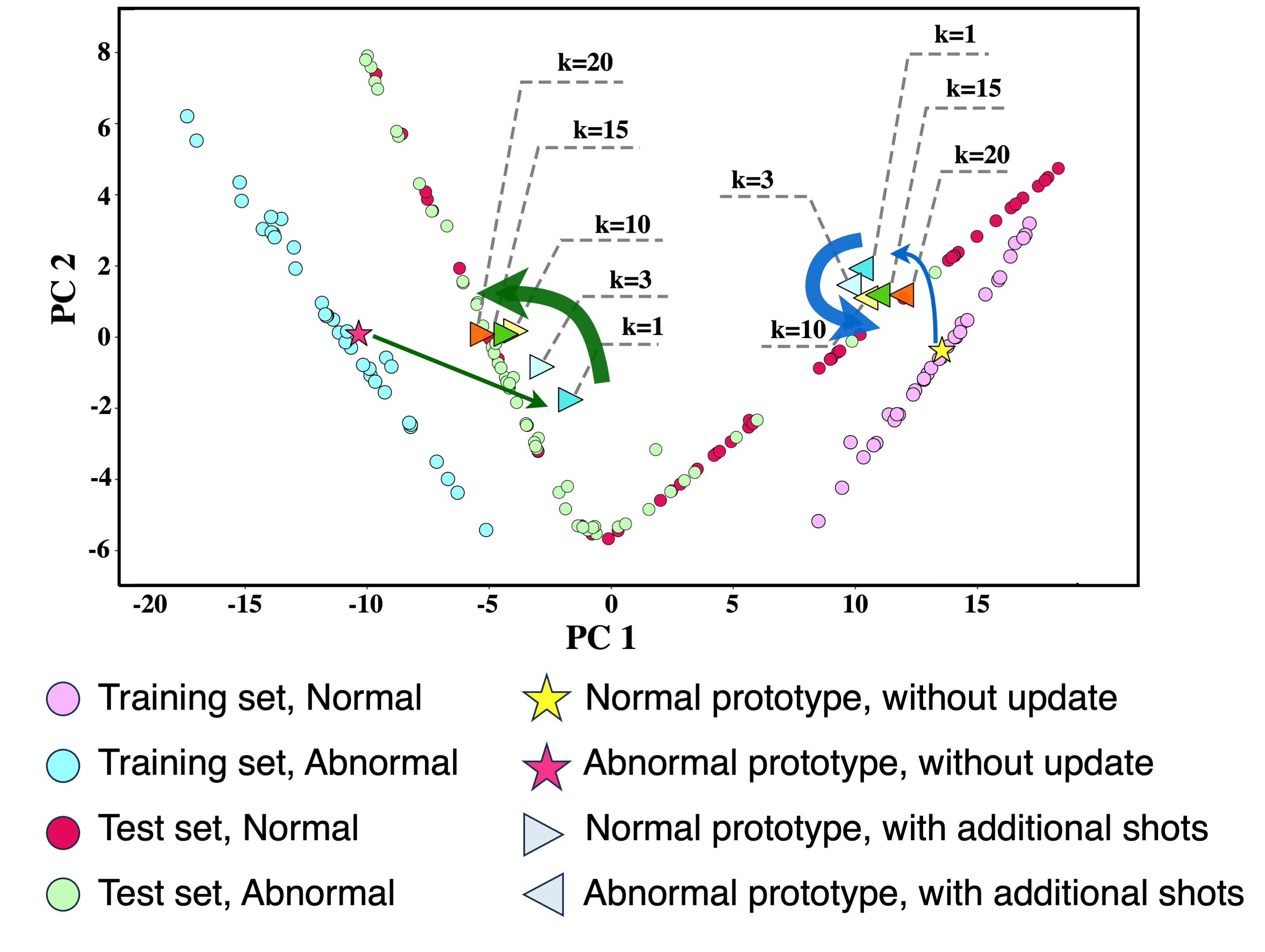}
     % \vspace{-1em}
    \caption{Prototype updating and their trajectory based on the number of new few shots to address Challenge \protect\circled{2}. As more new shots are involved,  the prototypes progressively align closer with the training and test data, effectively fine-tuning the model. The arrows in the figure show the movement of prototypes from their initial positions (marked by stars) toward better generalization with the test data. }
    \label{fig:pca_phase2}
\end{figure}

In Fig.~\ref{fig:pca_phase1}, we present a visual representation of the support and query data, highlighting the impact of fine-tuning in our meta-learning framework in Challenge~\circled{1}. Fig.~\ref{fig:pca_phase1}a presents the support and query data related to epilepsy patients, which has been subject to dimensionality reduction through Principal Component Analysis (PCA). In Fig.~\ref{fig:pca_phase1}b, we provide a parallel analysis for AF, with both cases being processed similarly using PCA for feature space transformation.
This visualization showcases the distribution of query data points after fine-tuning. In particular, our results reveal distinct clustering patterns between the support and query data after fine-tuning, indicating a substantial improvement in their separability. 

Furthermore, we show the results corresponding to Challenge~\circled{2}, where the data requirement for the process to update the values poses a significant consideration. To gain a better understanding of this phenomenon, we visualized the feature set of labeled samples in the target dataset and the samples in the test set in a two-dimensional representation using Principal Component Analysis~(PCA). Fig.~\ref{fig:pca_phase2} shows this representation for AF detection. The prototypes of the model, before receiving any update, are denoted by stars for each class. The arrows show the trajectory of the prototypes as they receive new shots from additional samples. 
As seen in Fig.~\ref{fig:pca_phase2}, when a single shot is received, the updated prototypes slightly deviate from the train and test samples. However, as more shots are received and the prototypes are updated with these shots, the prototypes align more closely with the test sample line. This result shows that with the additional shots and the updates of the prototypes, we can better generalize the fine-tuned model to the test set.

\subsection{Ablation study}

\begin{figure}[t]
    \centering
    \includegraphics[width=\linewidth]{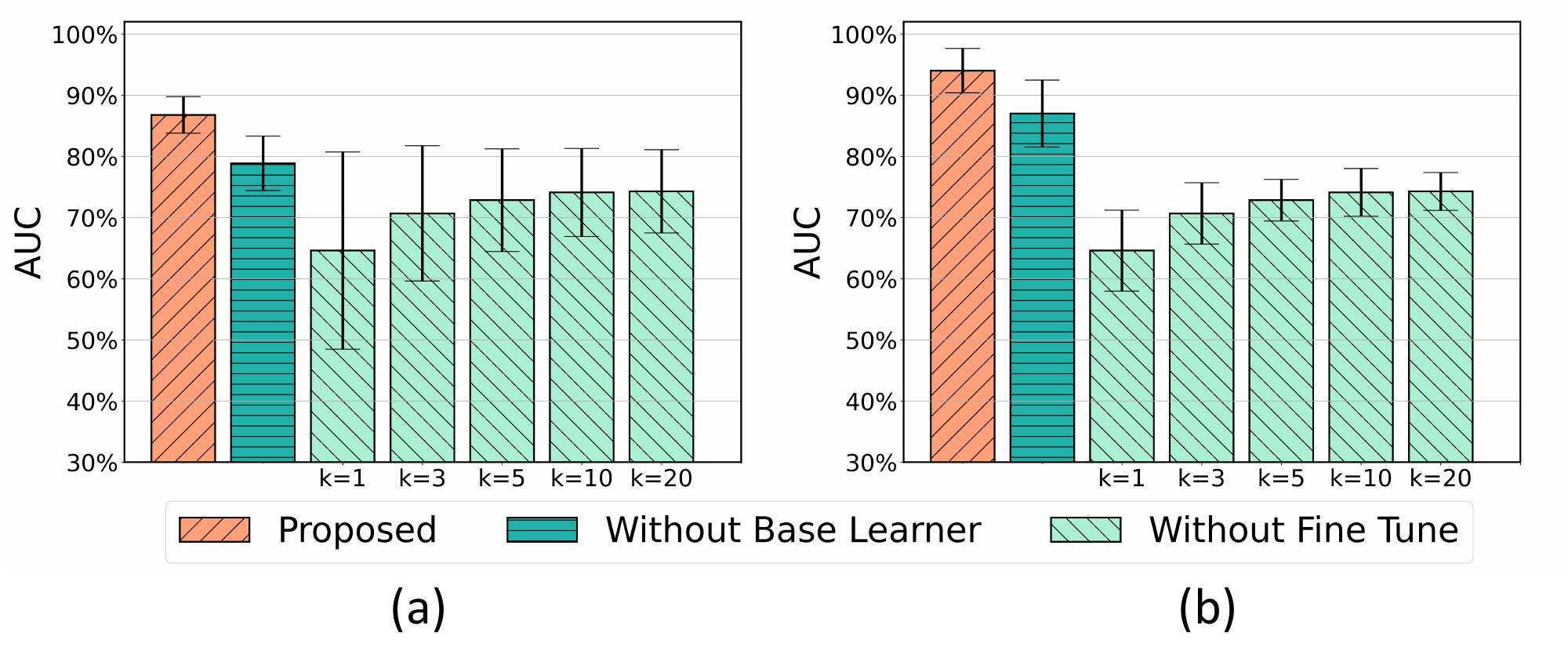}
    \caption{Comparison of the results for the two ablation study cases ``Without Fine-tune" and ``Without Base Learner" and the proposed method \ours{} for (a) epilepsy and (b) AF, showing the results for different shots 1, 3, 5, 10 and 20 in the case ``Without fine tune".}
    \label{fig:ablation}
\end{figure}

In order to investigate the effect of the base dataset, we conducted experiments without pretraining the model on the base dataset. 
% Our objective was to assess the model's performance when fine-tuned and tested with a limited sample of our dataset without the benefit of a comprehensive base dataset. 
According to the results shown in  Fig.~\ref{fig:ablation} for both AF and Epilepsy, we observed that the model's performance is significantly affected~(p-value $<$ 0.05) by the absence of a base dataset.
In particular, as shown in Fig.~\ref{fig:ablation}a for epilepsy, the AUC values exhibited an 8\% decrease without the use of the base dataset. In Fig.~\ref{fig:ablation}b, for AF, AUC is decreased by up to 7\%, demonstrating the advantages of pretraining the model on a base dataset in \ours{} to address Challenge~\circled{1}.

Further ablation study explores the effect of directly testing the target dataset without fine-tuning. In this configuration, we removed the fine-tuning step from the meta-training.
In  Fig.~\ref{fig:ablation}, we compared the results of this experiment with the results of our proposed strategy. For AF, the mean AUC shows a significant 10\% percent decrease (p-value $<$ 0.05), underscoring the critical role of fine-tuning in enhancing performance. Meanwhile, in the case of epilepsy, we observed a more pronounced effect, with a significant decrease in AUC of 12.5\% (p-value $<$ 0.05).

% This experiment emphasizes the pivotal role of fine-tuning in optimizing our model's performance and effectiveness for both AF and epilepsy, effectively demonstrating the substantial performance gains achievable through the fine-tuning process. These experimental setups provide a comprehensive view of our approach, shedding light on how different training and testing configurations impact model performance.

\subsection{Study limitations and future perspective}
\label{sec:limitation}
Our work presents two key limitations. Firstly, we trained \ours{} using publicly available datasets (TUH for seizure detection {\cite{obeid2016temple}} and PhysioNet for AF detection {\cite{clifford2017af}}) where data originated from hospital settings. However, the scarcity of public datasets involving wearable devices and data collected outside hospitals hinders a more generalizable evaluation. To partially address this, we employed the Siena dataset {\cite{detti2020siena}}, albeit small and still hospital-based, to simulate data from a different clinical setting. For AF detection, we were able to leverage a private dataset with wearable device data. The availability of publicly accessible wearable device datasets in the future holds promise for a more comprehensive evaluation of the effectiveness of our framework.

Secondly, the limited size of the target wearable device datasets restricted our ability to analyze the impact of seizure/AF types or the patients' age on our framework's performance. We believe that future research with larger datasets that separate patient data by AF/seizure types or ages would enable better fine-tuning in Challenge \circled{1} and targeted model updates focused on specific types of abnormalities in Challenge \circled{2}.

